\documentclass[10pt,a4paper]{article}

\usepackage{microtype}
\usepackage{graphicx}
\usepackage{multirow}
\usepackage{makecell}
\usepackage{booktabs} \usepackage{paralist}
\usepackage{xspace}

\usepackage{enumerate}
\usepackage[utf8]{inputenc} \usepackage[T1]{fontenc}    \usepackage{url}            \usepackage{booktabs}       \usepackage{amsfonts}       \usepackage{amsmath}       \usepackage{nicefrac}       \usepackage{microtype}      \usepackage{algorithmicx}
\usepackage{algorithm}\usepackage{algpseudocode}

\usepackage{mwe}    \usepackage{float}
\usepackage{lineno}
\usepackage{caption}
\usepackage[skip=0cm,list=true]{subcaption}

\usepackage{xcolor}
\definecolor{codegreen}{rgb}{0,0.6,0}
\definecolor{codegray}{rgb}{0.5,0.5,0.5}
\definecolor{codepurple}{rgb}{0.58,0,0.82}
\definecolor{backcolour}{rgb}{0.95,0.95,0.92}

\newcommand{\smartparagraph}[1]{\noindent{\bf #1}\ }

\renewcommand{\H}{\textbf{H}\xspace}
\renewcommand{\DH}{\textbf{DH}\xspace}
\newcommand{\BH}{\textbf{BH}\xspace}
\newcommand{\UNI}{\textbf{U}\xspace}

\usepackage[capitalise]{cleveref}
\usepackage{paralist}

\usepackage{authblk}

\begin{document}

\title{On the Impact of Device and Behavioral Heterogeneity in Federated Learning}

\author[1]{Ahmed M. Abdelmoniem}
\author[1]{Chen-Yu Ho}
\author[1]{Pantelis Papageorgiou}
\author[2]{Muhammad Bilal\thanks{Work done during an internship at KAUST.}}
\author[1]{ Marco Canini}
\affil[1]{KAUST}
\affil[2]{UCLouvain and IST(ULisboa)/INESC-ID}

\date{}

\maketitle

\begin{abstract}
Federated learning (FL) is becoming a popular paradigm for collaborative learning over distributed, private datasets owned by non-trusting entities. FL has seen successful deployment in production environments, and it has been adopted in services such as virtual keyboards, auto-completion, item recommendation, and several IoT applications. However, FL comes with the challenge of performing training over largely heterogeneous datasets, devices, and networks that are out of the control of the centralized FL server. Motivated by this inherent setting, we make a first step towards characterizing the impact of device and behavioral heterogeneity on the trained model. We conduct an extensive empirical study spanning close to 1.5K unique configurations on five popular FL benchmarks. Our analysis shows that these sources of heterogeneity have a major impact on both model performance and fairness, thus sheds light on the importance of considering heterogeneity in FL system design.
\end{abstract}

\section{Introduction}
\label{sec:intro}

The growing computational power of end-user devices (e.g., mobile phones)
coupled with the availability of rich sets of distributed data and concerns over transmitting private information make offloading model training to these devices increasingly attractive.
Federated Learning (FL)~\cite{mcmahan2017} is a machine learning approach wherein client devices owning different sets of data collaborate with the assistance of a central FL server to learn a global model without ever transmitting the private data.
Recently, FL has gained popularity as a range of practical applications and systems are readying deployment~\cite{yang2018applied, Bonawitz19, Li2020, Yang2019}.

A major challenge with FL, however, is dealing with the intrinsic heterogeneity of real-world environments.
To start with, data is typically Non-IID (independent
and identically distributed) across devices.
Thus, data, popularity and observation biases can creep into the trained model.
Besides, because FL is a participatory process, the quality of the resultant model depends on which devices partake in training, what training data is available, and how the runtime dynamics (e.g., client or network failures) have an effect. 

\begin{figure}[t!]
    \centering
    \includegraphics[width=0.75\linewidth]{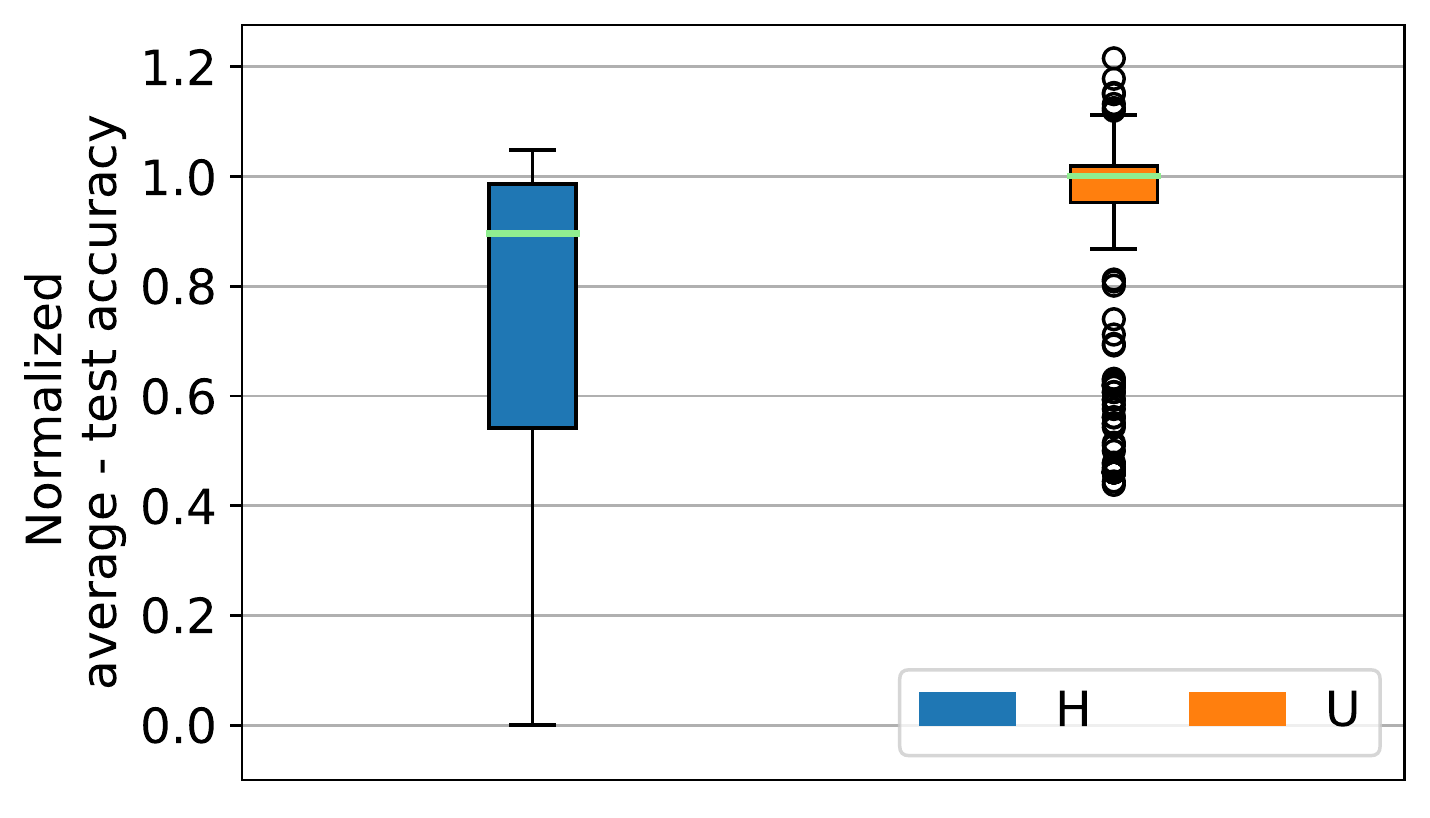}
    \caption{Impact of heterogeneity on model performance. The test accuracy is averaged over all devices and normalized by the baseline accuracy of each benchmark in its default configuration.
}
    \label{fig:relative_acc_boxplot}
\end{figure}

To appreciate the extent of the problem, Figure~\ref{fig:relative_acc_boxplot} contrasts two scenarios: (\textbf{U}) the ideal case where devices have uniform computational and network access capabilities which allow all of them to finish within the deadline and because they remain available throughout training, they can be sampled from uniformly at random; and (\textbf{H}) the realistic case where client devices have heterogeneous hardware and network access characteristics while at the same time client availability varies over time. 
The box-plot figure illustrates the test accuracy across a large number of experiments sweeping several hyper-parameters for five FL benchmarks (c.f. \S\ref{sec:experiments} for details).
The figure shows that heterogeneity has a significant impact on model performance: for 50\% of the experiments, the average accuracy is below 0.88$\times$ relative to corresponding baselines, and at the extreme, training does not converge.

The FL setting is riddled with different causes of heterogeneity, which may 
introduce bias in the trained model.
We focus in particular on \emph{device} and \emph{behavioral} heterogeneity.\footnote{Prior works~\cite{Mohri2019AgnosticFL,Chai2019,Sattler2019,Hard2020,Wang2020,Li2020,yang2020heter} have extensively studied and proposed mitigation methods for data heterogeneity. In contrast, device and behaviour heterogeneity are unexplored sources of performance degradation in federated learning.} 

Device heterogeneity (\textbf{DH}) is caused by variations in client devices, including hardware specifications, computational power, and the quality as well as the speed of Internet connectivity. \DH results in population bias in the training process due to unequal contribution by and/or failures of client devices. For instance, low-end devices or devices with unreliable network connections are more susceptible to failures or missing the server's reporting deadline.

Behavioral heterogeneity (\textbf{BH}) is caused by the churn in client devices.
Typically, FL systems consider a device to be available if its status is at the same time idle, plugged to a power source and connected to an unmetered network such as Wi-Fi~\cite{Bonawitz19}.
\BH results in sampling bias in the training process due to uneven participation of devices and unpredictable drop outs.

Despite a growing body of work in FL (see~\cite{kairouz2019advances} for a detailed survey), we find that surprisingly little attention has been given to the role of heterogeneity. Popular FL frameworks like TFF~\cite{tff} and Leaf~\cite{caldas2018leaf} do not directly allow studying the effects of DH and BH. Flash~\cite{yang2020heterogeneityaware} is the first work to develop a FL framework that simulates various sources of heterogeneity out of the box.
Building atop this framework, the focus of our work is to expose, characterize and quantify the effects of heterogeneity in the FL training process.
In particular, we answer the following questions: 
\begin{enumerate}
    \item {\em \textbf{Whether and how can heterogeneity impact the model performance and introduce bias in the FL training process?}}
    \item {\em \textbf{How sensitive is the FL training process to the choice of learning algorithm and FL hyper-parameters?}}
    \item {\em \textbf{Are existing proposals for addressing heterogeneity effective in mitigating sources of heterogeneity or are new methods necessary?}}
\end{enumerate}

To this end, we conduct an extensive set of experiments to collect a comprehensive set of measurements for analyzing empirically the impact of heterogeneity on model performance and fairness along with a systematic characterization of the main influencing factors. Our experiments span $\approx\!1.5K$ different configurations for five benchmarks resulting in a total of $\approx\!5K$ experiments and total compute time of $\approx\!36K$ GPU hours.

Our experimental results show that the impact of heterogeneity varies significantly, and, in many cases, it can lead to the divergence of the model (i.e., lack of convergence). Moreover, the \H scenario, which exhibits both device and behavioral heterogeneity, dominates the degradation in average performance and fairness in all scenarios. We also find that, in most cases, device heterogeneity (\DH) results in worse average performance, while behavioral heterogeneity (\BH) results in worse fairness in the model. Among all converged runs, we observe a clear correlation between heterogeneity and the performance and level of fairness in the model, leading to a degradation in both the fairness and average accuracy by up to 2.2$\times$ and 4.6$\times$, respectively. In other cases, heterogeneity leads to the divergence of the model, rendering the learning process useless. We also observe that heterogeneity's impact is quite sensitive to the choice of some hyper-parameters more than others. Overall, we believe our study and observations provide important and timely insights for FL researchers and system designers.

\section{Background}
\label{sec:background}

We review the FL paradigm with emphasis on its system design aspects. Then, we highlight how the learning algorithm and system configuration open up the possibility that environment heterogeneity may impact the trained model.

\subsection{Federated Learning}

\begin{figure}[!ht]
    \centering
    \includegraphics[width=1\linewidth]{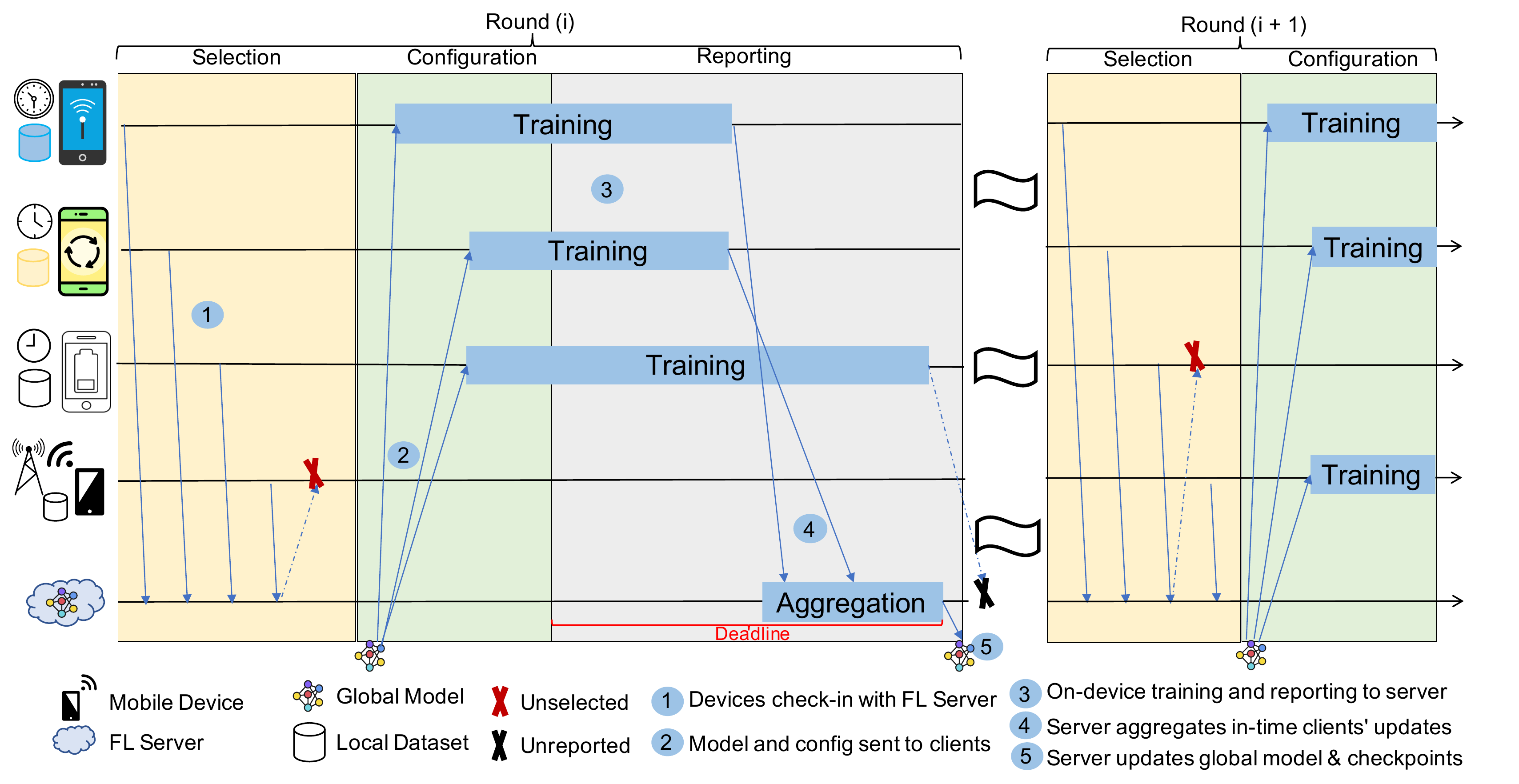}
    \caption{Phases of a federated learning system in heterogeneous setting.}
    \label{fig:federatedlearn}
\end{figure}

We focus on the popular FL paradigm introduced by \cite{mcmahan2017}, called federated averaging (FedAvg).
In this architecture, the learners (or clients) are end-user devices such as smartphones; training data is owned and stored at these devices; the learners train a global model collaboratively with the assistance of a centralized FL (or aggregation) server. The following may characterize the clients who participate in the training: 
\begin{inparaenum}
    \item they may have variable data points in number, type and distribution (\textit{data heterogeneity});
    \item they may use, to join the training, devices of wide variety of hardware configurations and network connections (\textit{device heterogeneity});
    \item They may not be available all the time due to reasons dependent on the end-user such as the device is not idle, not connected to WIFI or not connected to power source (\textit{behavioral heterogeneity}).
\end{inparaenum}
A common assumption is that the learners are honest whereas the server is honest but curious~\cite{Keith2018}. Hence, the training data always remain local to each learner and any leakage of information when communicating with the server should be avoided.

As depicted in \cref{fig:federatedlearn}, training of the global model occurs over a series of rounds, until the model converges to a satisfactory accuracy.
At the start of each round, the server waits for available devices to check-in. The server selects a subset of these devices which meet certain conditions such as being idle, and connected to WIFI and a power source. Then, the server sends the current version of the global model along with the necessary configurations (i.e., hyper-parameter settings) to the selected clients.
Each learner performs an equal number of local optimization steps as controlled by the \emph{Local Epochs} hyper-parameter set at the server. Then, learners send their updated models (or a model update, i.e., the delta from the global model) to the server. Typically, all communications are encrypted and some work proposed using differential privacy to prevent information leakage~\cite{Abadi2016}.
Finally, the server aggregates, with the global model, the model updates sent by the clients  (possibly via secure aggregation~\cite{Keith2018}) and then checkpoints the new global model to the local storage.

From a system perspective, the FL server -- typically realized as a scalable, distributed service -- oversees the entire training process following a predefined \emph{configuration}. For a given FL application, there might be tens to hundreds of thousands of available devices that are in communication with the server. Beyond maintaining the global model, as shown in \cref{fig:federatedlearn}, the server performs three phases that allow FL to be practical at scale: selection, configuration and reporting management, which are described below~\cite{Bonawitz19}.

\smartparagraph{Selection:} When each round starts, the server first waits for available devices to check-in. This takes place within the \emph{Selection Time Window}. From the set of connected clients, the server samples up to \emph{Selection Count} devices that will perform the training. If there are fewer devices by the end of the time window, the server progresses when at least there are \emph{Min Selection Count} devices; else it aborts the round.

\smartparagraph{Configuration:} After sufficient number of clients are selected, the server proceeds with sending the current version of the global model and the hyper-parameters of the training to the selected clients. Then, the clients train the received model on their local datasets using the hyper-parameters configuration pre-set by the server (e.g., number of epochs, batch size, learning rate, optimizer, etc).

\smartparagraph{Reporting:} After pushing the training procedure to the selected devices, the server waits for the devices to push their updates. The server uses a \emph{Reporting Deadline} as the timeout. The round completes successfully if, by the end of the deadline, at least \emph{Target Update Fraction} of devices report to the server; otherwise, the round fails and the received model updates are ignored.

The \emph{hyper-parameters} described above define a FL server configuration and need to be tuned for every FL application.

\subsection{Effects of Heterogeneity in Federated Learning}

Real FL deployments are exposed to a variety of environmental factors, like differences in data samples, device capabilities, quality of network links and availability (see \cref{fig:federatedlearn}). As a result, heterogeneity is endemic although its ultimate impact is not directly clear.
For instance, the popular FedAvg algorithm~\cite{mcmahan2017} assumes homogeneous devices, with equal probability of participating in training. In practice, \DH and \BH skew the distribution of participating devices.

From a system perspective, slower devices and devices with poor connectivity are less likely to meet the reporting deadline. To be less intrusive to the owners of devices, FL systems consider devices to be eligible when they are plugged to a power source, are connected to an unmetered network and are otherwise idle -- all factors influenced by user behavior.
Further, there are practical challenges regarding how to correct for these factors: in general, the FL server should not make use of any privacy-sensitive or user-identifying information~\cite{Bonawitz19}. This means that stateful solutions to compensate for heterogeneity are unlikely to be widely deployed.

Intuitively, device and behavioral heterogeneity induce population and sampling bias during training. Such a bias confounds with the underlying data bias (as data is Non-I.I.D. in general), and ultimately can negatively impact the trained model, which itself might present several indicators of bias:

\begin{itemize}
    \item \textbf{Lack of Fairness}: The model performance might be uneven across groups of users.
    \item \textbf{Lack of Robustness}: The model predictions could be inconsistent for different groups of users despite the input instances being largely similar across the groups.
    \item \textbf{Lack of Privacy}: The biased training may intensify the features or characteristics of overly-represented users and memorization of data samples may lead to leakage of private data.
\end{itemize}

We view our work as a first step towards characterizing the effects of heterogeneity-induced bias. To make our study concrete and tractable, among the above issues, we focus on the fairness aspect.
Ideally, FL should ensure that the model is fair to all groups of participating users, under some definition of fairness. The existence of bias can be revealed by measuring the level of fairness among the participants.

We use Jain's fairness index~\cite{jain1999throughput} to measure the level of fairness in a distribution of values. Jain's index is commonly used as a measure of fairness of the attained throughput among TCP flows that compete for scarce network bandwidth. 
Jain's fairness index $F_I$ is expressed as:
\begin{equation}
\textstyle    F_I = \frac{(\sum_{i=1}^{N}x_i)^2}{N \times \sum_{i=1}^{N}x_i^2},
\end{equation}
where, $N$ is the number of clients and $x_i$ is the per-client performance measure under consideration for fairness evaluation (typically for us, the clients' test accuracy).

\section{Methodology}
\label{sec:methodology}
To characterize and quantify the impact of heterogeneity on model performance and fairness, we use an experimental design approach following these driving questions:

\begin{enumerate}[Q1)]
    \item Is there a definite trade-off between model performance/fairness and heterogeneity-induced bias?
    \item To what extent does behavioral versus device heterogeneity affect the bias and how do their individual effects confound?
    \item How sensitive is model performance/fairness to the configuration of FL-specific hyper-parameters?
    \item Does the choice of the optimizer (SGD, SGD with momentum, or Adam) have an influence on the impact of heterogeneity?
    \item How do application-specific hyper-parameters (like learning rate, batch size, or number of epochs) influence the impact of heterogeneity?
    \item Are recently proposed FL algorithms designed for heterogeneous settings (i.e., FedProx~\cite{Li2020FedProx} and Q-FFL~\cite{Li2020Fair}) effective in mitigating the bias resulting from device and behavioral heterogeneity? How sensitive are these algorithms to influencing environment (e.g., composition of devices) or configuration (e.g., reporting deadline) factors?
\end{enumerate}

\subsection{Experimental Design} 

To answer the above questions, we follow the experimental design approach. ``Experimental Design'' refers to the process of executing controlled experiments in order to collect information about a specific process or system~\cite{the-design-of-experiments}. The experimental design approach comprises a careful selection of influencing factors chosen to allow an accurate view of the system's response. Finally, repeating runs allows to account for the variance of individual experiments.

\smartparagraph{Factors:}
The factors are the heterogeneity setting (or lack thereof) and the many hyper-parameters that influence the FL process. We subdivide them into three categories: environment, application-specific, and FL system/algorithm. We list them out in \cref{tab:fl-hyperparamters,tab:algo-hyperparamters,tab:agg-hyperparamters} along with the ranges of experimented values.

\smartparagraph{Experiments:}
The space of possible instantiations of factors is huge. Besides, as with hyper-parameter tuning in general machine learning, many choices of values might not be valid, which makes it hard to use a space-filling approach to cover the large experimental space uniformly.

To principally cover the space of experiments, we identify a default configuration for each benchmark and we then systematically perform experiments while varying factors, typically one by one as a deviation from the default configuration.
However, certain characteristics of any experiment are random: e.g., the assignment of data partitions to devices,  the proportion of different type of devices, or the available clients during selection phase.
We control for variance by repeating experiments five times using distinct seeds to initialize randomness.

A noteworthy aspect of a benchmark's default configuration is the reporting deadline. We search for the critical deadline, an appropriate minimal value of this hyper-parameter such that the fraction of successful clients over the rounds is on average above the target update fraction.

\smartparagraph{Platform:}
We use Flash~\cite{yang2020heterogeneityaware} to run realistic experiments. Flash simulates runs of FL applications and faithfully models wall-clock execution time while multiplexing execution of many devices onto a single GPU.
Flash also comes with a trace of user behavior collected from a real-world FL application deployed across three countries.
Flash bins devices into three capability-classes: low-end (LE), moderate (M) and high-end (HE) devices. The computational speed of these groups follow the execution profiles of three real-world devices~\cite{yang2020heterogeneityaware}. Finally, the devices' network access links in terms of upload and download speeds, are randomly chosen from 20 different distributions covering a wide-variety of network conditions observed in practice.

\smartparagraph{Heterogeneity:}
Since our objective is to tease out the influence of heterogeneous settings, we consider four scenarios for every experiment.
The baseline case (\UNI) is the ideal scenario where clients are always available and uniform (i.e., their devices are homogeneous in hardware and link speed) and the server sets a large enough deadline for all clients to finish in time. The heterogeneity scenarios, are three: 
\begin{inparaenum}
    \item device heterogeneity (\DH) - the clients are always online but their device hard and link speed are sampled at random from the real-world trace and link speed distributions, respectively;
    \item behavioral heterogeneity (\BH) - the clients use the moderate device model and same link speed but their availability follows the timeline of the user in the real-world trace~\cite{yang2020heterogeneityaware};
    \item full heterogeneity (\H) - is the simultaneous combination of (\DH) and (\BH) where device model and link speed are sampled at random and client availability follows the real-world user trace. 
\end{inparaenum}

\begin{figure*}[!h]
  \centering
    \includegraphics[width=1\linewidth]{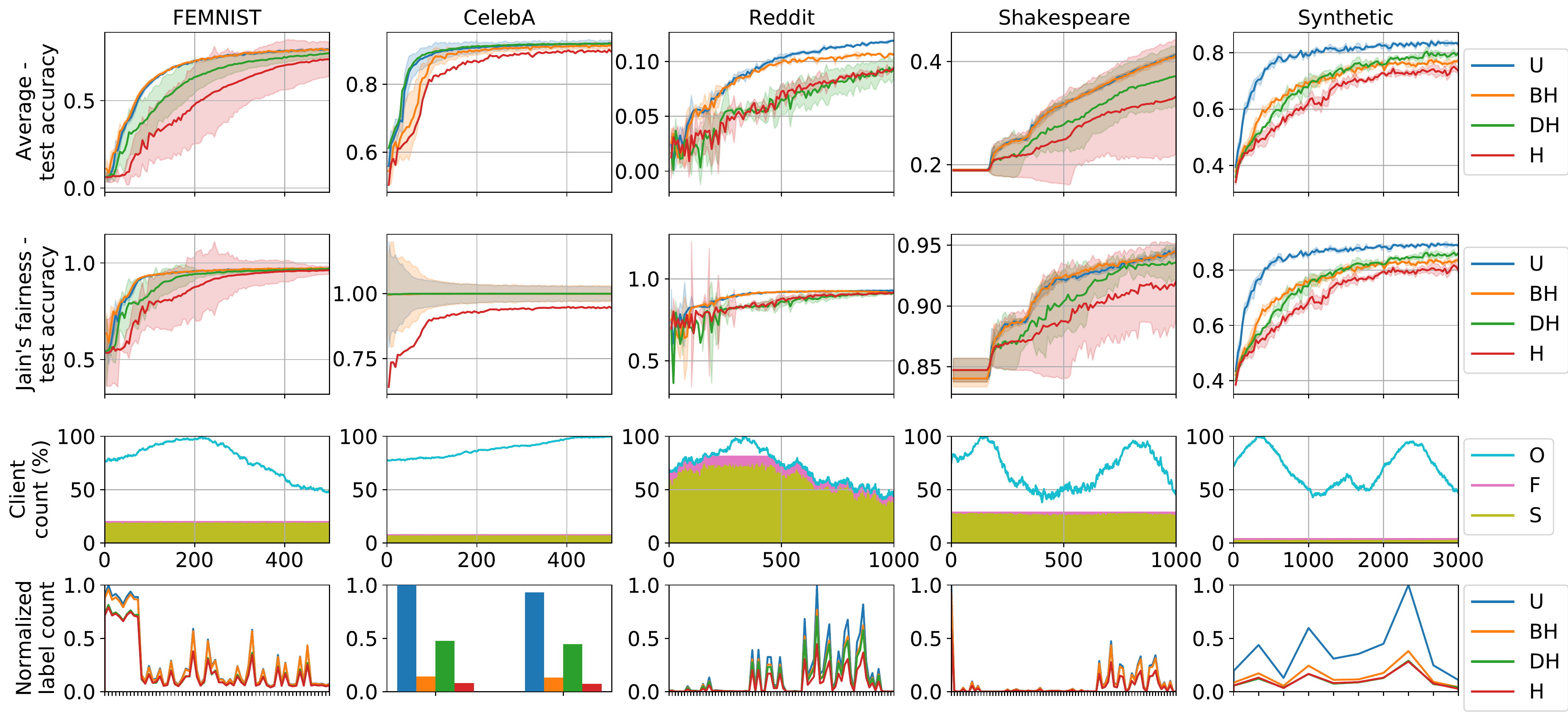}
    \caption{Impact of different heterogeneity settings on model performance and level of fairness for different FL benchmarks. The $1^{st}$ and $2^{nd}$ rows show the median value (solid line) and one standard deviation (filled range) among all seeds for the average test accuracy and Jain's fairness index, respectively. The $3^{rd}$ row shows the online (O), failed (F) and successful (S) client counts throughout rounds. The $4^{th}$ row shows normalized frequency for each label on which the clients trained throughout training. Note that, CelebA (binary classification) has only 2 labels, thus using grouped bars. X-axis label for rows 1, 2, and 3 is round number and is unique data labels for row 4.}
    
    \label{fig:heter-impact}
\end{figure*}

\begin{table*}[!t]
\caption{Summary of the benchmarks used in this work.}
\resizebox{\linewidth}{!}{\begin{tabular}{ccccccccccc}
    \toprule
        Task & \makecell{ML \\ technique} & Model 	&	Dataset	& \makecell{Model \\ Size [bytes]} & \makecell{Total \\ Clients} 
        &  \makecell{Selection \\ Count \\ (\textbf{sc})} & \makecell{Reporting \\ Deadline \\ (\textbf{ddl}) $[s]$} & \makecell{Maximum \\ Sample Count} & \makecell{Learning\\Rate} & \makecell{Quality \\ metric}   
        \\\midrule
        \multirow{2}{*}{\makecell{Image \\ Classification}} & \multirow{2}{*}{CNN} & ~2 Conv2D Layers \ & FEMNIST~\cite{EMNIST}	& 26,414,840 & 3,400  
        & 100  & 60 & 340 & 0.01  & 79.91\%
        \\ 
        & & 1 Conv2D Layer & CelebA~\cite{CELEBA} &	124,808  & 9,343  & 100 &  15 & 30  & 0.01 & 90.63\%
        \\ 
        \midrule
       
         \multirow{2}{*}{\makecell{Next Word \\ Prediction}} &  \multirow{2}{*}{RNN} & \multirow{2}{*}{LSTM} & Reddit~\cite{REDDIT} & 24,722,496 & 813  
         & 100 & 27  & 50 & 0.5 & 11.88\%
         \\
          &  & & Shakespeare~\cite{mcmahan2017} & 3,271,488  & 1,129
          & 50 & 142 & 50  & 0.8 & 40.10\%
         \\ 
         \midrule
         
         \makecell{Cluster \\ Identification} &  \makecell{Traditional \\ ML} &  \makecell{Logistic \\ Regression} & Synthetic~\cite{caldas2018leaf} &	2,400 & 9,367 
        & 50  & 23  & 340 & 0.005 & 83.00\%
         \\
        \bottomrule
    \end{tabular}
    }
    \label{tab:benchmarks}
\end{table*}

\section{Experimental Evaluation}
\label{sec:experiments}

\begin{figure}[t]
  \centering
    \includegraphics[width=0.8\linewidth]{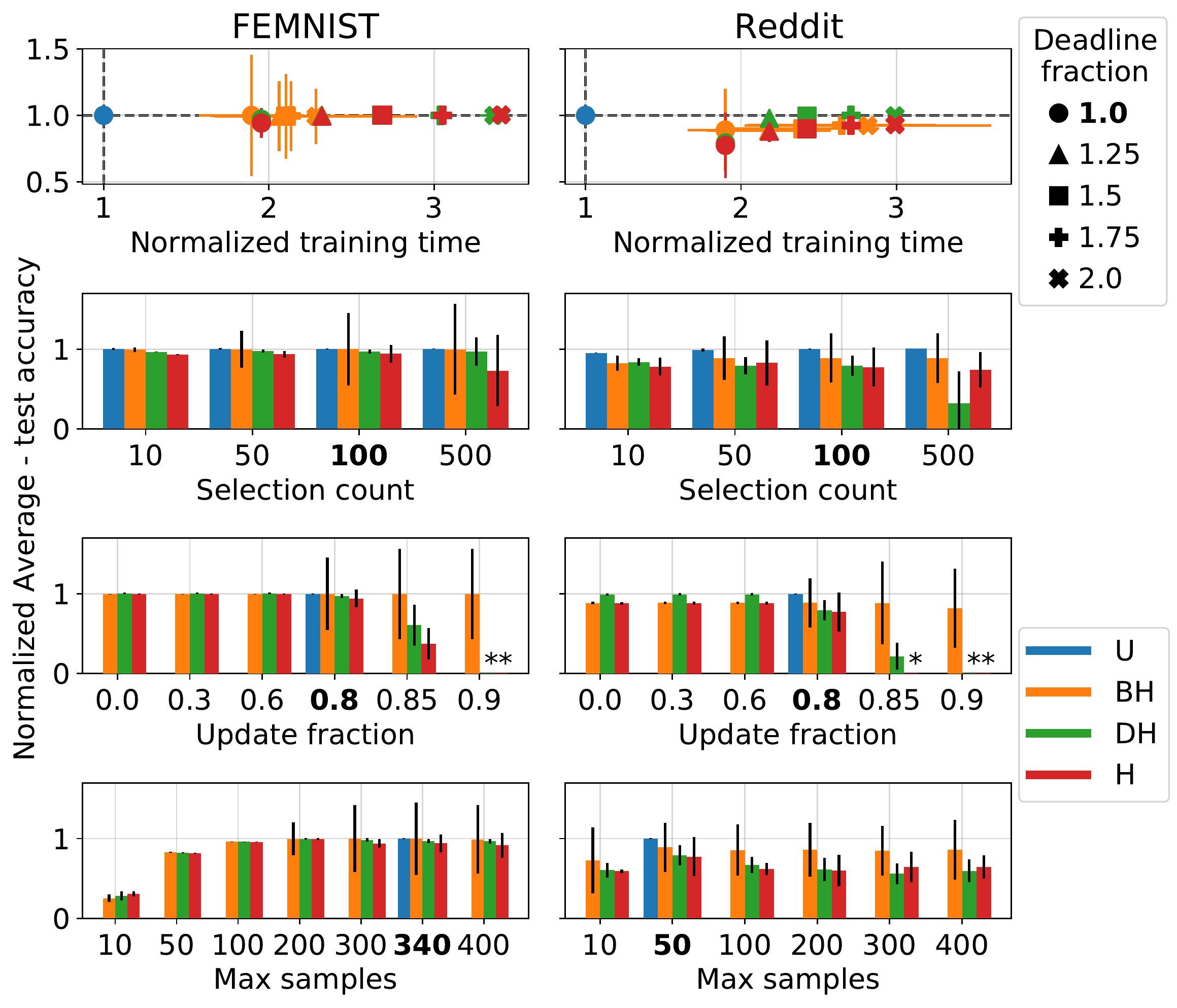}
    \caption{The sensitivity of the average accuracy to the choice of the FL hyper-parameters for FEMNIST and Reddit Benchmarks. We mark the default settings of the uniform scenario in bold in all following figures.}
    \label{fig:avg-test-acc-FL-params}
\end{figure}

\begin{figure*}[!ht]
  \centering
    \includegraphics[width=1\linewidth]{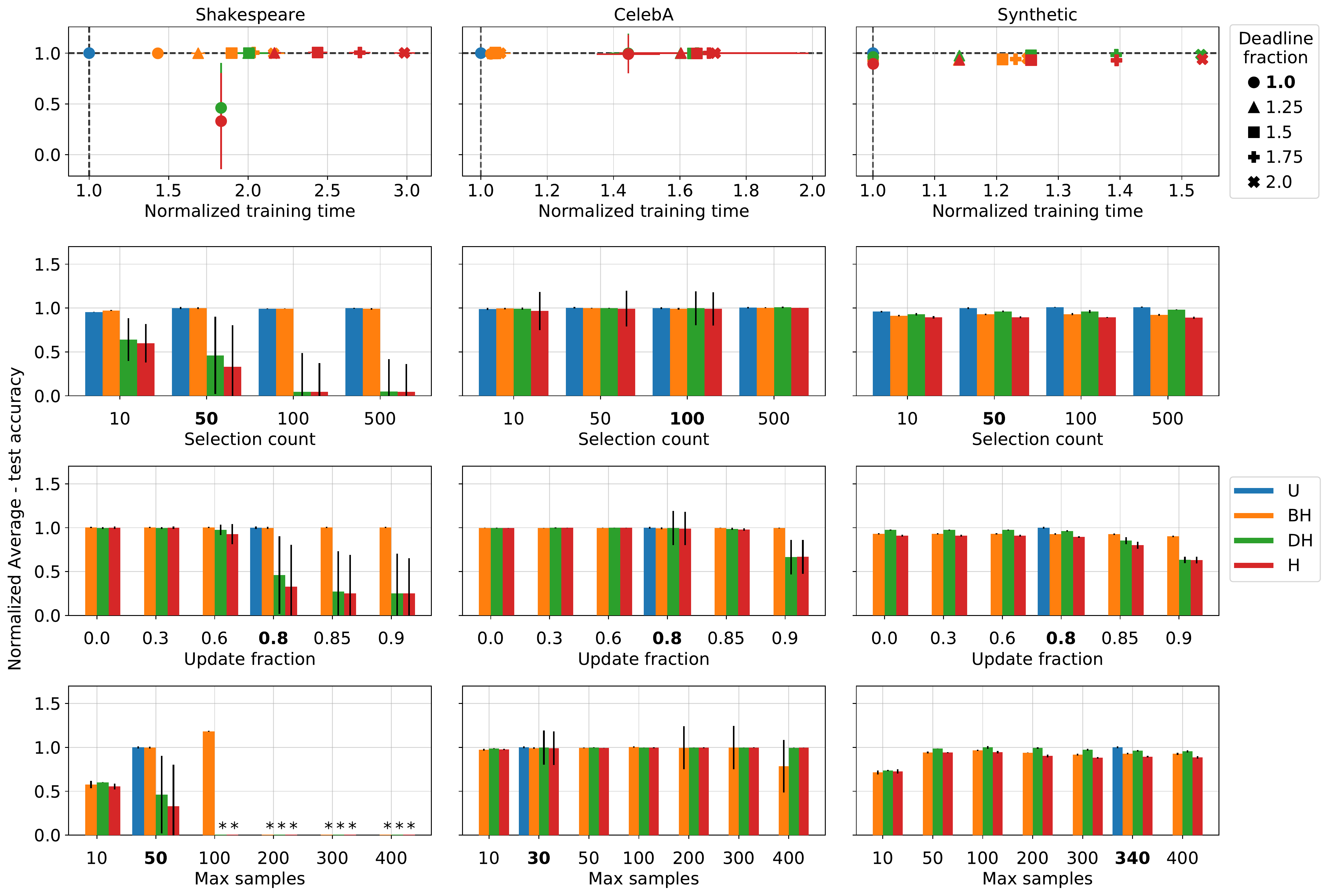}
    \caption{Sensitivity of the FL hyper-parameters choice for Shakespeare, CelebA and Synthetic Benchmarks.}
    \label{fig:app:avg-test-acc-FL-params-apdx}
\end{figure*}

\smartparagraph{Benchmarks:} 
We use five benchmarks covering a variety of FL applications used in several prior works~\cite{mcmahan2017,Li2020Fair,Li2020FedProx,yang2020heterogeneityaware}.

\cref{tab:benchmarks} summarizes the application, dataset, and default configuration of each benchmark. 
See \S\ref{apdx:datasets} for further details regarding the datasets.

We partition the training and testing datasets so that each client owns a partition of each dataset as done in prior works.
During a training round, a selected client uses samples from its training partition.
To evaluate the model accuracy on the testing dataset, we run the test rounds 100 times during the whole experiment (i.e., test rounds are set to run every $\frac{R}{100}$ training round where $R$ is the total training rounds). For each round, we evaluate the model accuracy at every client or from a random sample of 3,500 clients, if the total client count is higher than this cap.

The default configuration of each benchmark is primarily based on the information from prior works~\cite{hard2018federated,caldas2018leaf,Bonawitz19}. The reporting deadline is set by us based on the search of the critical deadline (\S\ref{sec:methodology}).
For all benchmarks, the server uses a selection time window of 20 seconds and the clients use SGD as their local optimizer. We also note that the default \emph{batch size} is 10, \emph{number of epochs} is 1, \emph{min selection count} is 10 and \emph{update fraction} is 0.8 (80\%). 
Finally, the default FL aggregation algorithm we use is FedAvg~\cite{mcmahan2017} as employed in~\cite{Bonawitz19} wherein only successful clients get their model updates reflected in the global model.

\begin{table}[!t]
\centering
\caption{FL hyper-parameters. \textbf{ddl} and \textbf{sc} are the default reporting deadline and selection count as in \cref{tab:benchmarks}.}
\resizebox{0.68\columnwidth}{!}{\begin{tabular}{ |c|c| } 
        \hline
        \textbf{Hyper-parameter} & \textbf{Values} \\
        \hline
        Selection count & [10, 50, 100, 500] \\ 
        \hline
        Max samples per client & [100, 200, 300, 400] \\ 
        \hline
        Target update fraction & [0.0, 0.3, 0.6, 0.8, 0.85, 0.9] \\ 
        \hline
        Deadline & [1.25$\times$\textbf{ddl}, 1.5$\times$\textbf{ddl}, 1.75$\times$\textbf{ddl}, 2$\times$\textbf{ddl}] \\
        \hline
        \multirow{7}{*}{(LE\%, M\%, HE\%)} & [(0.0, 0.0, 1.0), (0.0, 1.0, 0.0), \\ 
        & (1.0, 0.0, 0.0), (0.0, 0.5, 0.5), \\
        & (0.5, 0.5, 0.0), (0.5, 0.0, 0.5), \\
        & (0.1, 0.1, 0.8), (0.8, 0.1, 0.1), \\
        & (0.1, 0.8, 0.1), (0.2, 0.2, 0.6), \\
        & (0.2, 0.6, 0.2), (0.6, 0.2, 0.2), \\
        & (0.33, 0.34, 0.33)] \\
        \hline
        Min selection count & [0.25$\times$\textbf{sc}, 0.5$\times$\textbf{sc}, 0.75$\times$\textbf{sc}, \textbf{sc}] \\ 
        \hline
        \end{tabular}
}
\label{tab:fl-hyperparamters}
\end{table}

While our focus is on realistic conditions with Non-I.I.D. data, we also ran each benchmark using an I.I.D. version of the dataset. Our analysis of both versions support that the results and observations reported below for the Non-I.I.D. case are not due to data heterogeneity.

\smartparagraph{Setup:}
We run experiments using the Flash FL simulator on a GPU cluster.
We run $\approx\!5K$ experiments, requiring a total of $\approx\!36K$ GPU-hours ($\approx\!4$ years).

Next, we discuss the results in detail. We mainly present the average test accuracy and Jain's fairness index, both normalized by the corresponding values of the baseline (i.e., the setting (\UNI) with the default configurations). In the figures, the missing values, which are marked as (*), represent the divergent cases. That means that particular setting and hyper-parameter configuration, for all the repetitions of its experiment with different seeds, have diverged).

\subsection{Heterogeneous settings and their impact}

\cref{fig:heter-impact} presents the model performance (the average test accuracy across devices) and fairness (based on Jain's fairness index) for every benchmark contrasting the heterogeneous scenarios (\BH, \DH, \H) to uniform (\UNI).
We observe that both \DH and \BH have an impact, and generally, they together confound (\H) to yield even greater impact on both the model's performance and fairness.

The impact of \BH becomes more significant for benchmarks that either have a relatively small client population (Reddit) or have a small number of data samples per client (CelebA and Synthetic). These cases are more sensitive to variations in device availability as it affects the quantity and diversity of the training data. Over the converged runs, the degradation caused by \BH on average performance and fairness ranges from 1.1$\times$ to 3.9$\times$ and from 1$\times$ to 2.2$\times$, respectively.

The impact of \DH is in general more profound despite all devices are always available. This is because the device type is determined at random following a distribution skewed in favor of low-end devices. Since, the device's ability to finish training before the deadline depends on the time needed to process the device's data samples, low-end devices are more likely to fail to finish the training in time. In our \DH and \H experiments, there are on average 55\% LE, 44\% M, and 1\% HE devices. Over the converged runs, the degradation caused by \DH on average performance and fairness ranges from 1.13$\times$ to 4.1$\times$ and from 1$\times$ to 1.96$\times$, respectively. 

In general, on average, \BH results in more degradation to the fairness of the model, while \DH results in more degradation to the average performance. Therefore, the composition of device types and client availability is the main contributor to the average performance and fairness degradation seen for \H. This degradation can be attributed to the confounding effect of \DH, which results in a mixture of device types dominated by low-end devices and \BH, which results in the unavailability of moderate and high-end devices. As a result, within heterogeneous settings, the global model is updated using updates from fewer clients than the uniform case. Therefore, as shown via the label frequency in \cref{fig:heter-impact}, the model sees a lower number of training data samples and unique clients. This observation helps explain the degradation in both the average performance and fairness of the global model.

\smartparagraph{Takeaways:}
These observations illustrate the importance of taking into consideration the type and sources of heterogeneity when designing FL systems in practice. Therefore, FL system designers should be aware of the impact of heterogeneity in general and \DH in particular. The impact might be reduced by means of employing some form of proportional load balancing during the client selection phase~\cite{Nishio2019}, computation offloading techniques at the edge~\cite{Wu2020}, or asynchronous mode of model updates from the clients~\cite{Lu2020}.

\subsection{FL hyper-parameters and heterogeneity settings}

We now study the sensitivity of the heterogeneity settings to the choice of FL hyper-parameters.

We observe that the average performance is quite sensitive to the choice of the FL hyper-parameters as shown in \cref{fig:avg-test-acc-FL-params} and \cref{fig:app:avg-test-acc-FL-params-apdx}. We also observe similar trends with fairness as shown in \cref{fig:fairness-test-acc-FL-params} and \cref{fig:app:fairness-test-acc-FL-params-apdx} of the appendix.  With reference to the figures, we describe the effects of each hyper-parameter below.

\smartparagraph{Reporting Deadline:} The deadline is one of the key FL hyper-parameters, which directly influences the success rate of the clients. The default setting is set to allow, on average, just enough fraction of clients to submit the updates in time. The results show that increasing the deadline results in improvement of the obtained average test accuracy. Moreover, the improvements are more pronounced for \DH and \H settings. This is because a longer deadline allows enough time for more devices (which dominantly, 99\% of them, are low-end to moderate) to submit their updates in time. However, we observe that the improvements flatten at a certain point after which the extra time spent in training due to waiting for the deadline would be wasteful. Hence, there is a clear trade-off between the performance and run-time costs which can be mainly controlled by tuning the reporting deadline. 

\smartparagraph{Selection Count:} We observe that a higher selection client count leads to almost no improvement in the average model performance for \UNI and \BH settings. However, we observe larger variations, with larger numbers of selection client count, as indicated by the error bars for the \BH setting. In contrast, for \DH and \H settings, we observe that by increasing the selected client count, the model performance and fairness are severely degraded (esp. for the Shakespeare benchmark). This can be attributed to both the over-fitting of the model due to larger global batch sizes coupled with the increased number of client failures in \DH and \H in which the majority of clients are low-end devices. 

\smartparagraph{Target Update Fraction:} We note that, for all heterogeneity settings, the target update fraction of successful clients directly impacts both the performance and fairness and introduces a noticeable trade-off. Specifically, higher targets for update fraction results in a higher probability that the number of successful clients does not reach the target. This leads to an increase in the number of abandoned rounds and hence results in a lower level of fairness and average quality of the model (or divergence in many cases especially for \DH and \H). In contrast, a lower update fraction, while it ensures that fewer number of rounds would not abandoned, results in contributions from fewer clients towards model updates and hence increased model bias. Therefore, the tuning of this hyper-parameter is critical for the efficient operation of FL systems.

\smartparagraph{Maximum Samples:} is the maximum number of samples each client is allowed to use when participating in training or testing. This is useful for Non-IID data to bound each client's contributed samples which would make the learning updates more fair among clients. However, the results show lower values of maximum number of samples can limit the learning process (i.e., slow convergence and hence low model performance) which equally impacts all settings. In general, the results show that, in \BH settings, the quality and fairness of the model are mildly impacted by the choice of the maximum number of samples. In contrast, we observe noticeable impact (and in some cases divergence) for both \DH and \H settings. This is because, for large maximum number of samples, clients with low-end devices, needing to process more samples per epoch, will likely fail to report within the deadline. In addition, the maximum number of samples results in a trade-off between ensuring a higher degree of inclusion and diversity of clients' data. This suggests that it should be tuned depending on the distribution of data points available with the clients at the time of training.

\smartparagraph{Extent of Device Heterogeneity:} 
In realistic settings, the device type of the clients would be largely random. Even though, the default setting for \DH chooses the device types at random, we run experiments where we control the fraction of the device types assigned to the clients to quantify the extent of \DH's impact on both model performance and fairness. 

\begin{figure}[!t]
  \centering
    \begin{subfigure}{0.85\linewidth}
    \includegraphics[width=1\linewidth]{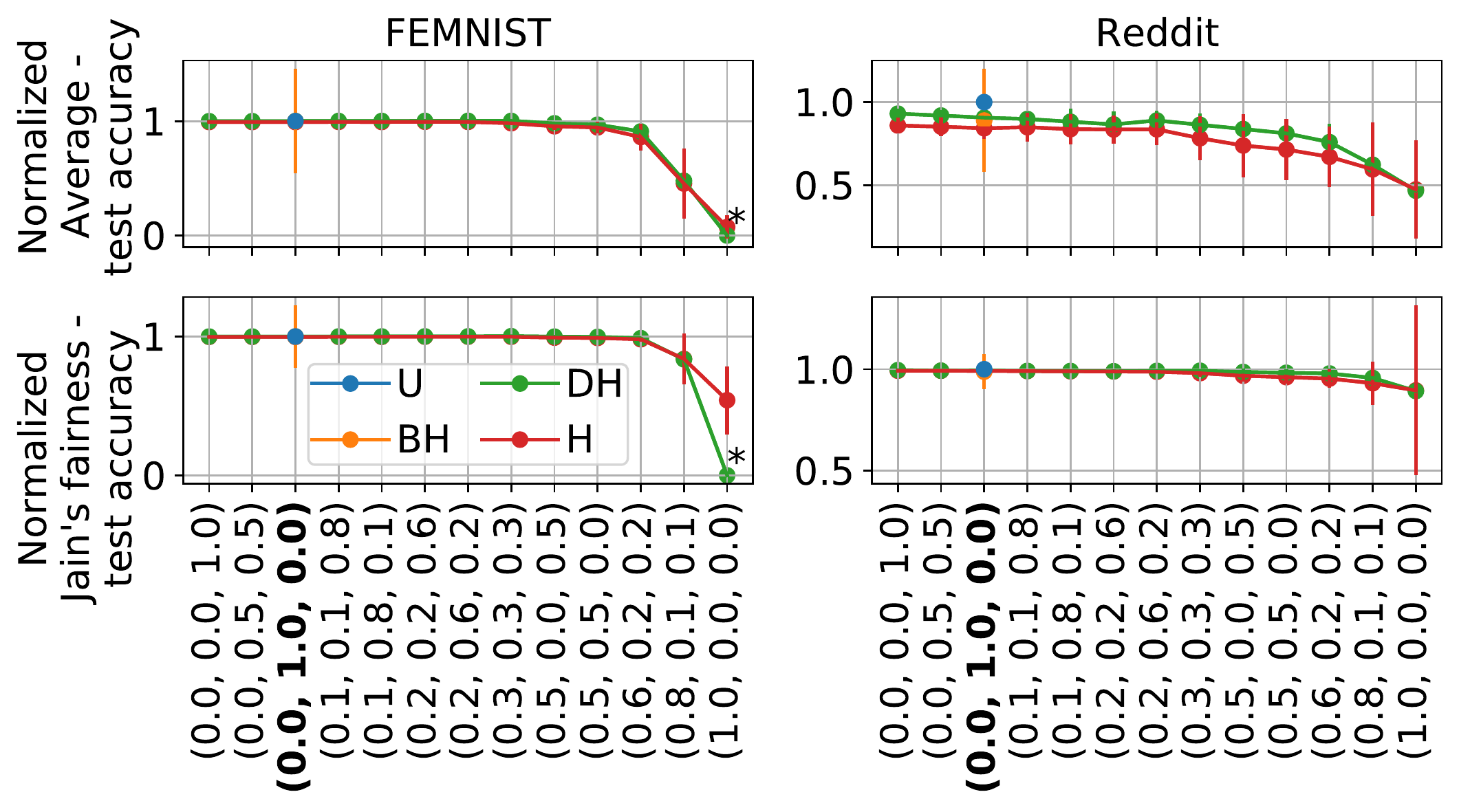}
     \caption{FEMNIST (Left) and Reddit (Right).}
    \end{subfigure}
    \begin{subfigure}{1\linewidth}
    \includegraphics[width=1\linewidth]{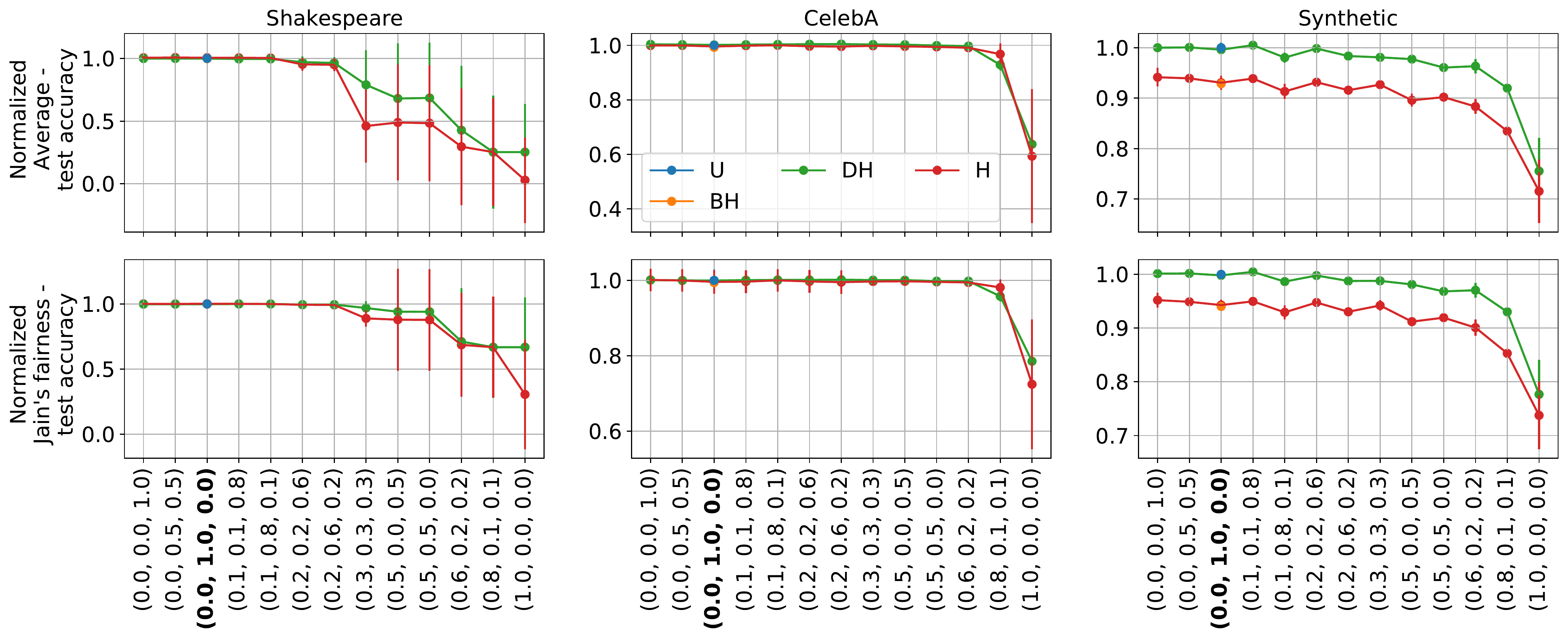}
    \caption{Shakespeare (Left), Celeb (Middle) and Synthetic (Right).}
    \end{subfigure}
    \caption{Impact of varying device fractions on model accuracy and levels of unfairness. For example, (0.2, 0.2, 0.6) means 20\% low-end, 20\% mid-end, and 60\% high-end devices. Normalized Test Accuracy and Jain's Fainress are shown on top and bottom rows, respectively.}
    \label{fig:dev-heter-impact}
\end{figure}

\cref{fig:dev-heter-impact} shows that, in all scenarios, both the average model accuracy and level of fairness remarkably decrease as more clients with low-end devices become part (or the majority) of the clients' population. These devices tend to not be able to finish training or uploading within the reporting deadline. We also observe that the impact are amplified to the extreme of resulting into divergence if the task is computationally heavy (e.g., larger model in FEMNIST or more samples to process in Shakespeare). These results suggest that training models without careful consideration to the computational capabilities and the type of network access of clients' devices can significantly result in biased and low quality models.

\begin{figure}[!t]
  \centering
    \includegraphics[width=0.8\linewidth]{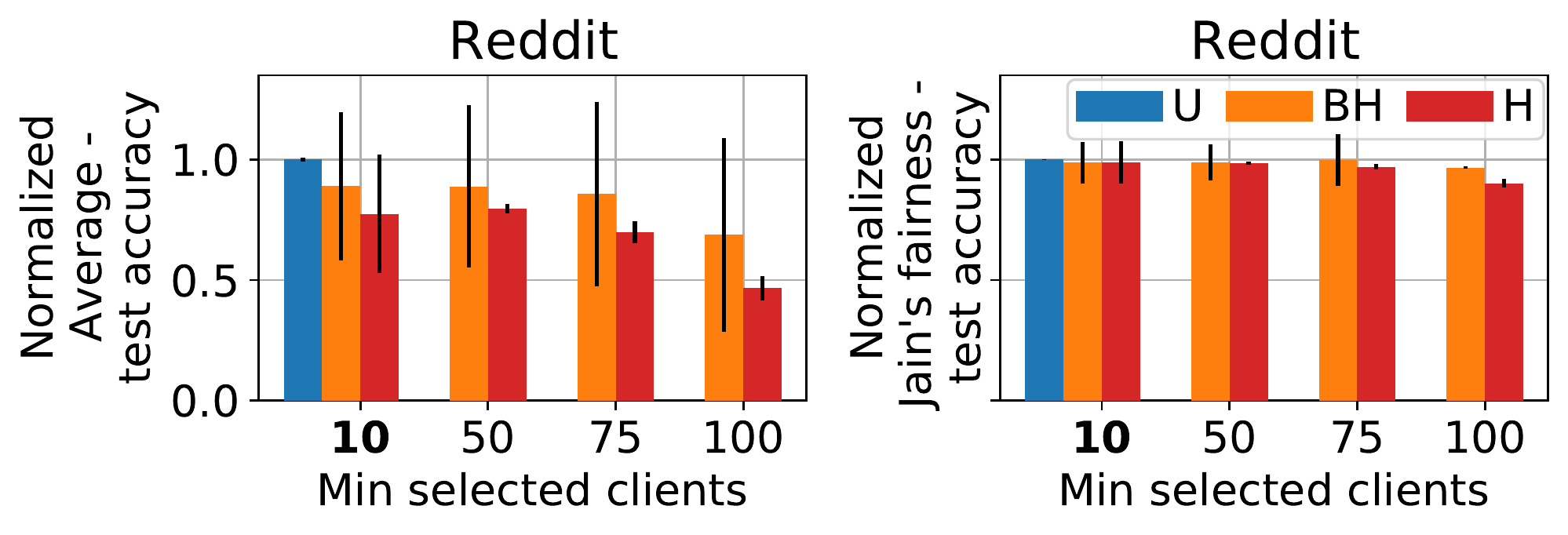}
    \caption{The impact of varying minimum selection which influences behavioral heterogeneity.}
    \label{fig:behv-heter-impact}
\end{figure}

\smartparagraph{Behavior Heterogeneity:} \emph{Min selection count} is one of the FL hyper-parameters that imposes a threshold on the minimum number of the clients selected from the online ones for which the server waits to be available or otherwise the round is abandoned. This hyper-parameter is only relevant to the selection stage and primarily is influenced by user behavior (\BH). This is because \BH results in variability in the number of online clients over training rounds. A min selection count ensures training only proceeds with enough devices to allow for reasonable quality model updates and limiting the bias towards smaller groups of clients. 

\cref{fig:behv-heter-impact} shows that, the more restrictive (i.e., higher number) the min selection count is, the higher the degradation in the average model performance. This is mainly because the higher the min selection count, the higher the chance of selection stage failures. This observation is more profound for the Reddit benchmark because it has low total client population and hence a smaller proportion of online clients as shown in \cref{fig:heter-impact}. Typically, a higher number of selected clients is preferred (e.g., around 120-130\% of the target count to account for any client failures or drop-outs). Therefore, the choice of minimum selection count introduces a trade-off between the probability of failing the selection stage and having enough clients to account for  failures.

\subsection{Importance of heterogeneous settings and hyper-parameters}
\begin{figure}[!htbp]
  \centering
    \includegraphics[width=0.7\linewidth]{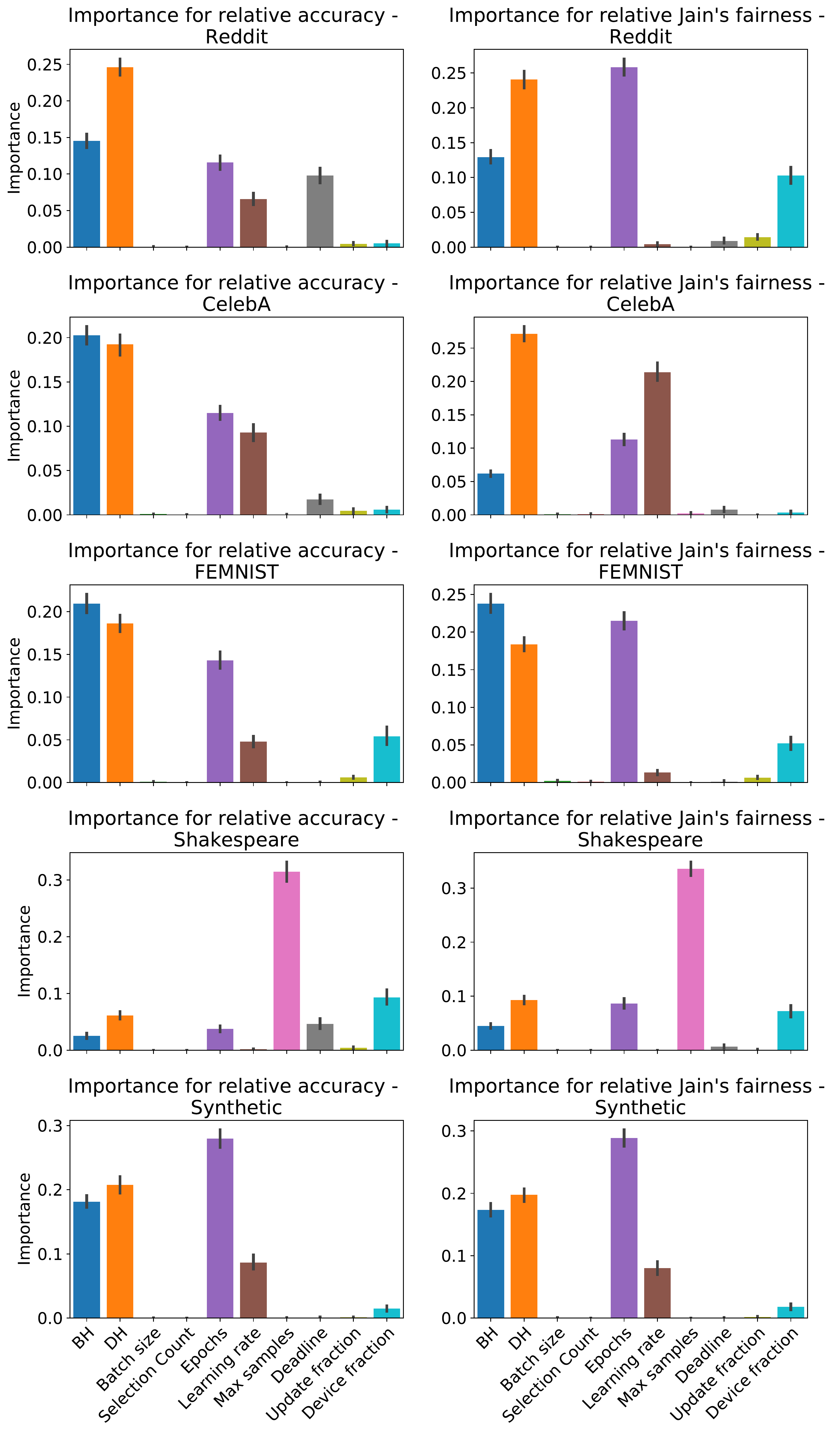}
    \caption{Functional ANOVA analysis.}
    \label{fig:fanova}
\end{figure}
We have seen how much the choice of hyper-parameters can impact the performance of the heterogeneity settings. Next, we identify how much each factor influences the performance variation using the performance metrics of interest (i.e., normalized average test accuracy and Jain's fairness). 

To this end, we leverage functional ANOVA analysis~\cite{fanova} to evaluate the importance of different factors towards the impact on model performance.
\cref{fig:fanova} shows the individual importance of different factors on the experiments. The error bars show the confidence interval around the estimated values based on 100 repeated runs of functional ANOVA. We consider only convergent runs for this analysis.
We see from the figure that, for most benchmarks, \BH, \DH and number of epochs are the most important factors. Hence, \BH and \DH have a significant impact on the model performance and fairness, while the remaining hyper-parameters have a smaller influence to the performance variations. The importance of number of epochs is linked to the default value for the reporting deadline: a larger number of epochs make it more likely to miss the deadline. We see similar trends in most of other benchmarks and for the situations when fairness is the performance metric of interest. In particular, for 7 out of 10 cases, \BH, \DH and number of epochs are the three most important hyper-parameters but their relative ranking varies in different cases. 

The exceptions of these observations are the Shakespeare benchmark (for both accuracy and fairness performance metric) and CelebA benchmark (only for fairness performance metric). For CelebA benchmark with Jain's fairness index, learning rate, \DH and number of epochs are the three most important hyper-parameters. For Shakespeare benchmark, users have many more samples that they can process and increasing the threshold on the maximum samples can cause clients to miss the reporting deadline due to increased computational needs. Therefore, max samples has the highest importance in that case.

\smartparagraph{Takeaways:} The aforementioned observations show the significance of careful tuning of the FL system hyper-parameters while taking into account the type heterogeneity it influences the most. Moreover, most of the hyper-parameters introduce trade-offs and practitioners should pay more attention when setting them and balance between objectives or give more weight for the most important ones. Moreover, FL system designers should be aware of \DH in particular and employ some form of load balancing among the participating devices.

These observations spark the intriguing question of what if some level of non-sensitive information about clients' device hardware and network were available to the server, could the impact of heterogeneity be mitigated. This might be achieved via means of adaptive per-client tuning of the reporting deadline, target update fraction and/or the maximum samples. For mitigation of \BH impacts, the server may keep temporal historical information on clients' participation trends and dynamically adapt the minimum selection count for each round based on this information.

\subsection{Optimizer and related hyper-parameters}
\begin{table}[!h]
\small
\caption{Learning optimizers and their hyper-parameters.}
\centering
\resizebox{0.5\columnwidth}{!}{\begin{tabular}{ |c|c| } 
        \hline
        \textbf{Hyper-parameter} & \textbf{Values} \\
        \hline
        Optimizer & [SGD, SGD-Momentum, Adam] \\ 
        \hline
        Learning Rate & [0.001, 0.01, 0.05, 0.1, 0.5] \\
        \hline
        Batch size & [5, 50, 100] \\
        \hline 
        Epochs & [5, 10, 20] \\
        \hline
        \end{tabular}
}
\label{tab:algo-hyperparamters}
\end{table}

We next explore whether the choice of the optimizer would result in an obvious variation in the performance or fairness for different settings. To this end, we run experiments using the optimizers listed in~\cref{tab:algo-hyperparamters}. Note, for Adam, we set the learning rate two orders of magnitude lesser than the one used for SGD~\cite{luo2018adaptive}. 

\begin{figure}[!t]
  \centering
    \begin{subfigure}{0.78\linewidth}
    \includegraphics[width=1\linewidth]{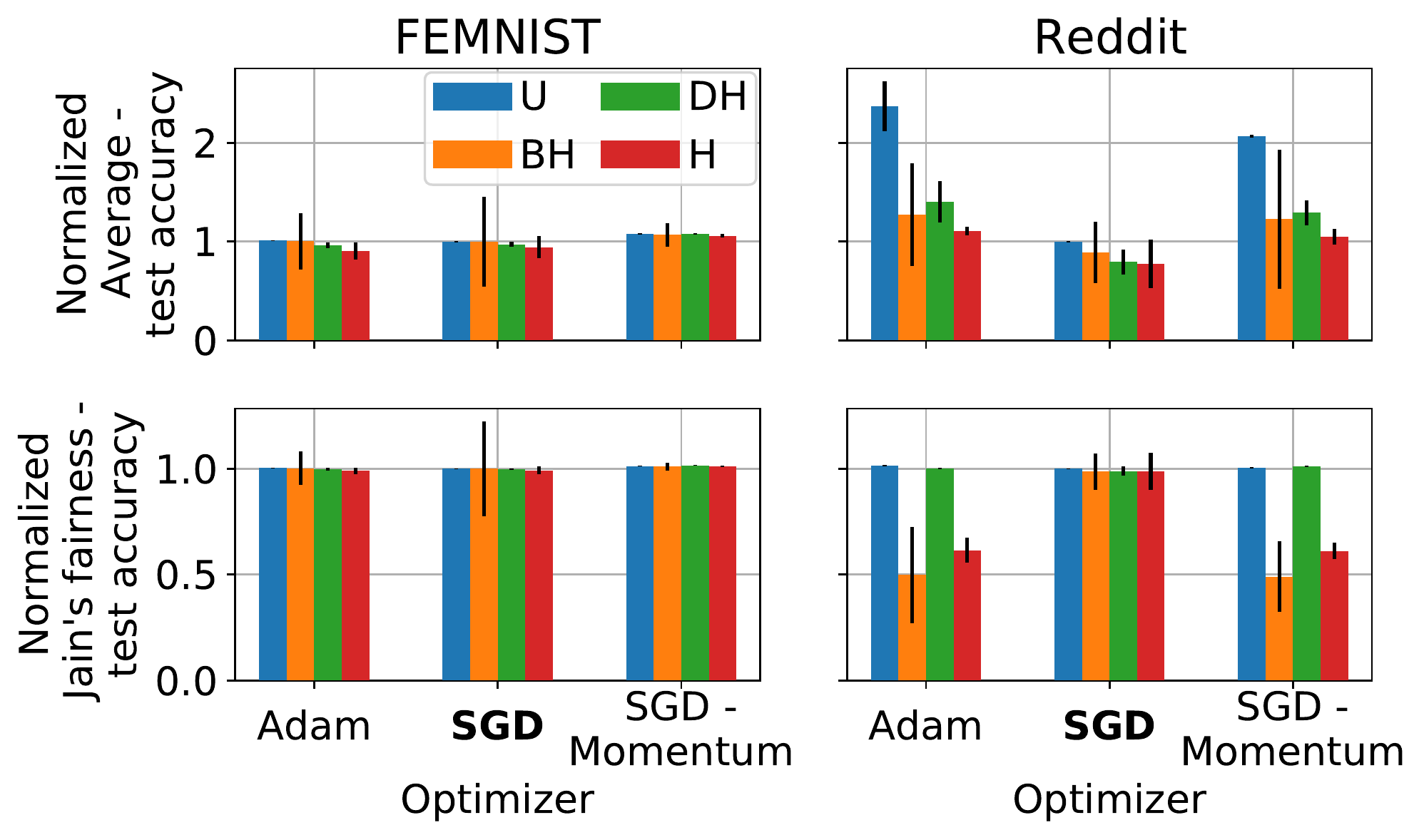}
     \caption{FEMNIST (Left) and Reddit (Left).}
    \end{subfigure}
    \begin{subfigure}{1\linewidth}
    \includegraphics[width=1\linewidth]{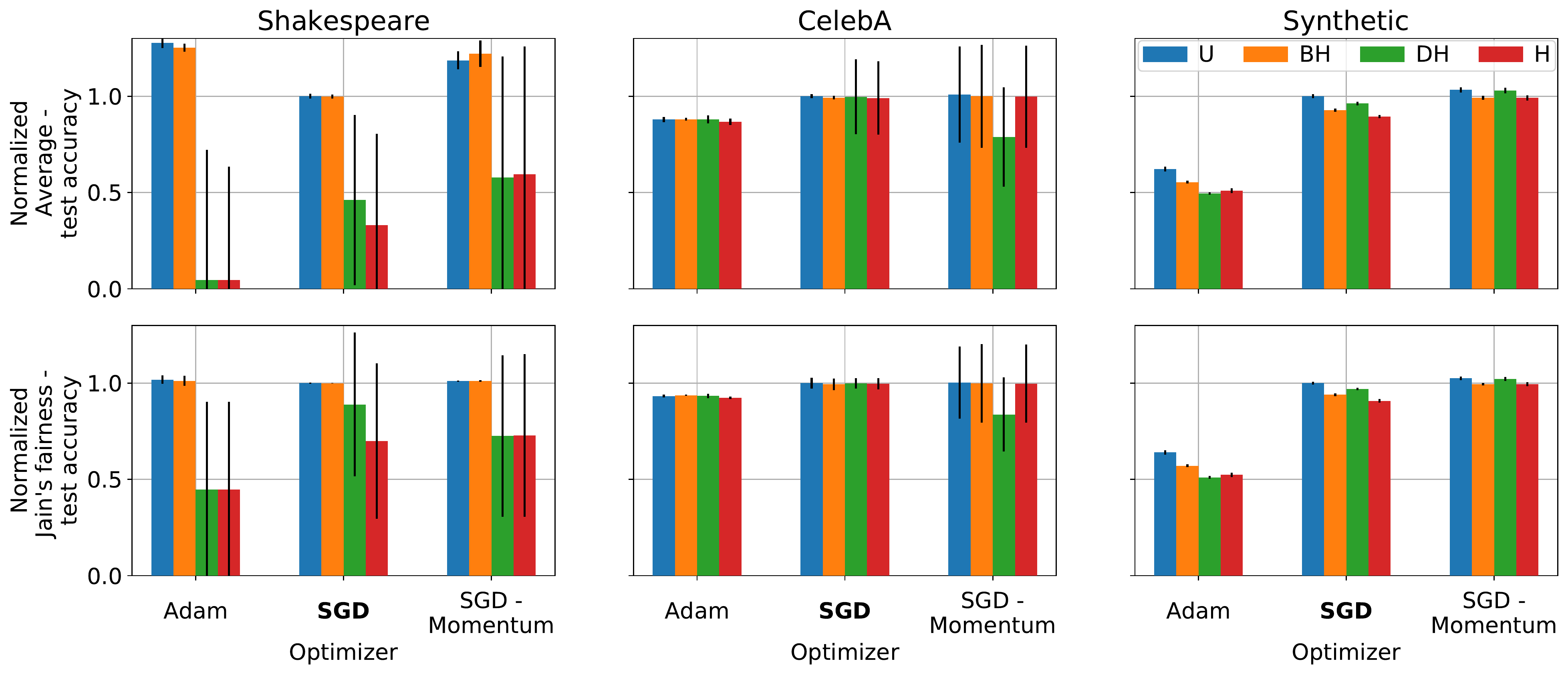}
     \caption{Shakespeare (Left), Celeb (Middle) and Synthetic (Right).}
    \end{subfigure}

    \caption{The impact of the optimizer on model performance and fairness. Top row is normalized average test accuracy and bottom row is normalized Jain's Fairness Index.}
    \label{fig:opt-heter-impact}
\end{figure}

\smartparagraph{Optimizer: } As depicted in \cref{fig:opt-heter-impact}, we observe slight changes in both the average performance and fairness of the trained model for FEMNIST. However, for Reddit, we observe significant improvement in average accuracy but with lower fairness when using Adam or SGD-Momentum. For other benchmarks, Adam results in noticeable reduction for the both performance and fairness (esp., Shakespeare, and Synthetic). These results indicate that the choice of optimizer is important as it can directly influence heterogeneity and its impact. However, the influence of the optimizers is largely dependent on the learning task and therefore should be chosen carefully with respect to the task. And, practitioners should be more cautious when using adaptive optimizers such as Adam in heterogeneous settings.

\smartparagraph{Takeaway:} In general, the choice of the optimizer is only important to the nature of the learning task but, in some cases, it may influence the effects of the heterogeneity in FL environments. Therefore, FL task designers should be careful when choosing the optimizer for their learning tasks and take into consideration its influence on the impact from heterogeneous settings.

Next, we evaluate the model's sensitivity to the variations in the choice of the application-specific hyper-parameters. The results for normalized average test accuracy are presented in \cref{fig:hyper-test-acc-train-params} while \cref{fig:hyper-fairness-test-acc-train-params} of the appendix shows the results for normalized Jain's fairness index of test accuracy.

\begin{figure}[!t]
  \centering
  \begin{subfigure}{0.78\linewidth}
   \includegraphics[width=1\linewidth]{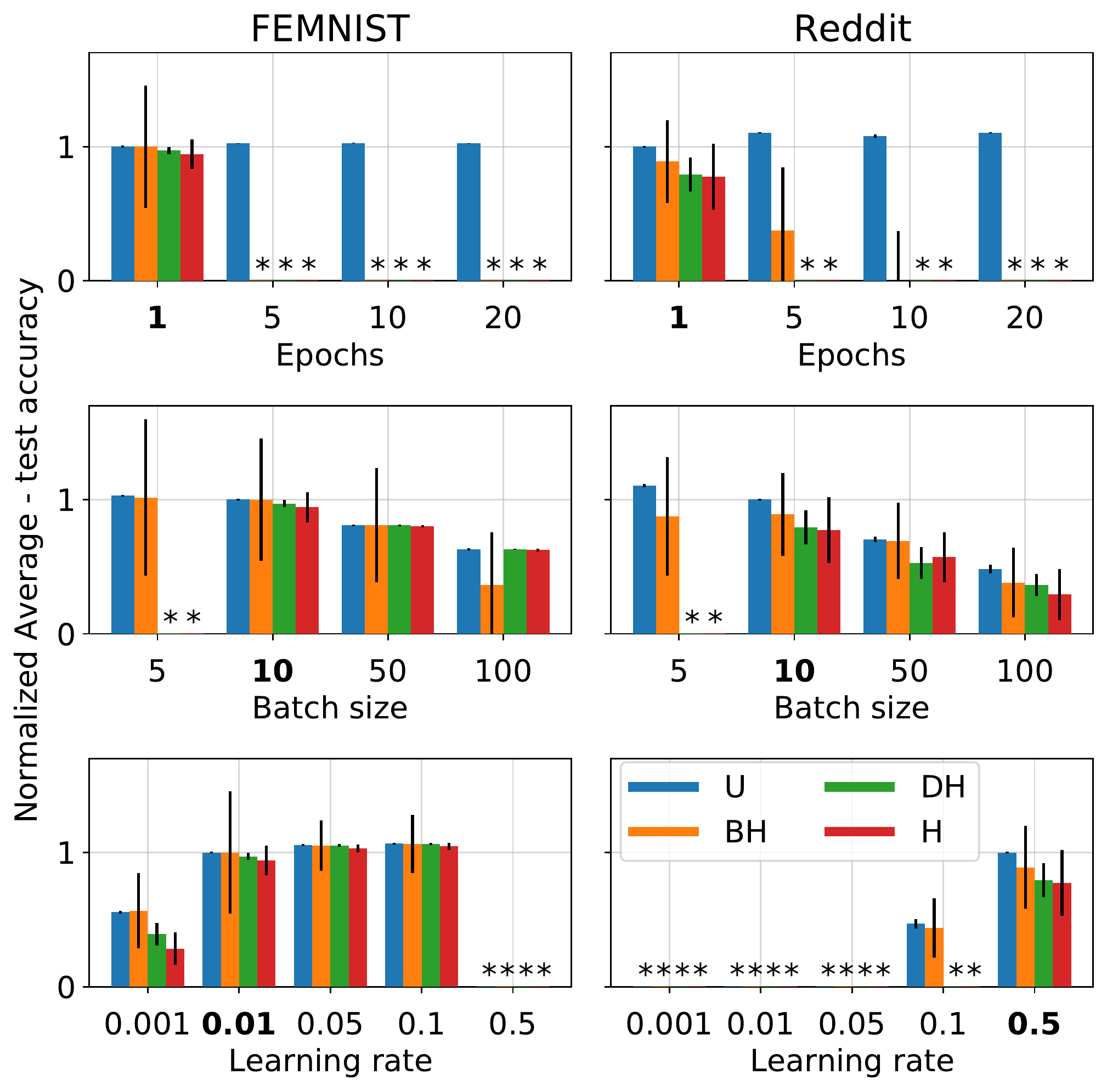}
   \caption{FEMNIST (Left), and Reddit (Right).}
  \end{subfigure}
  \\
    \begin{subfigure}{1\linewidth}
    \includegraphics[width=0.95\linewidth]{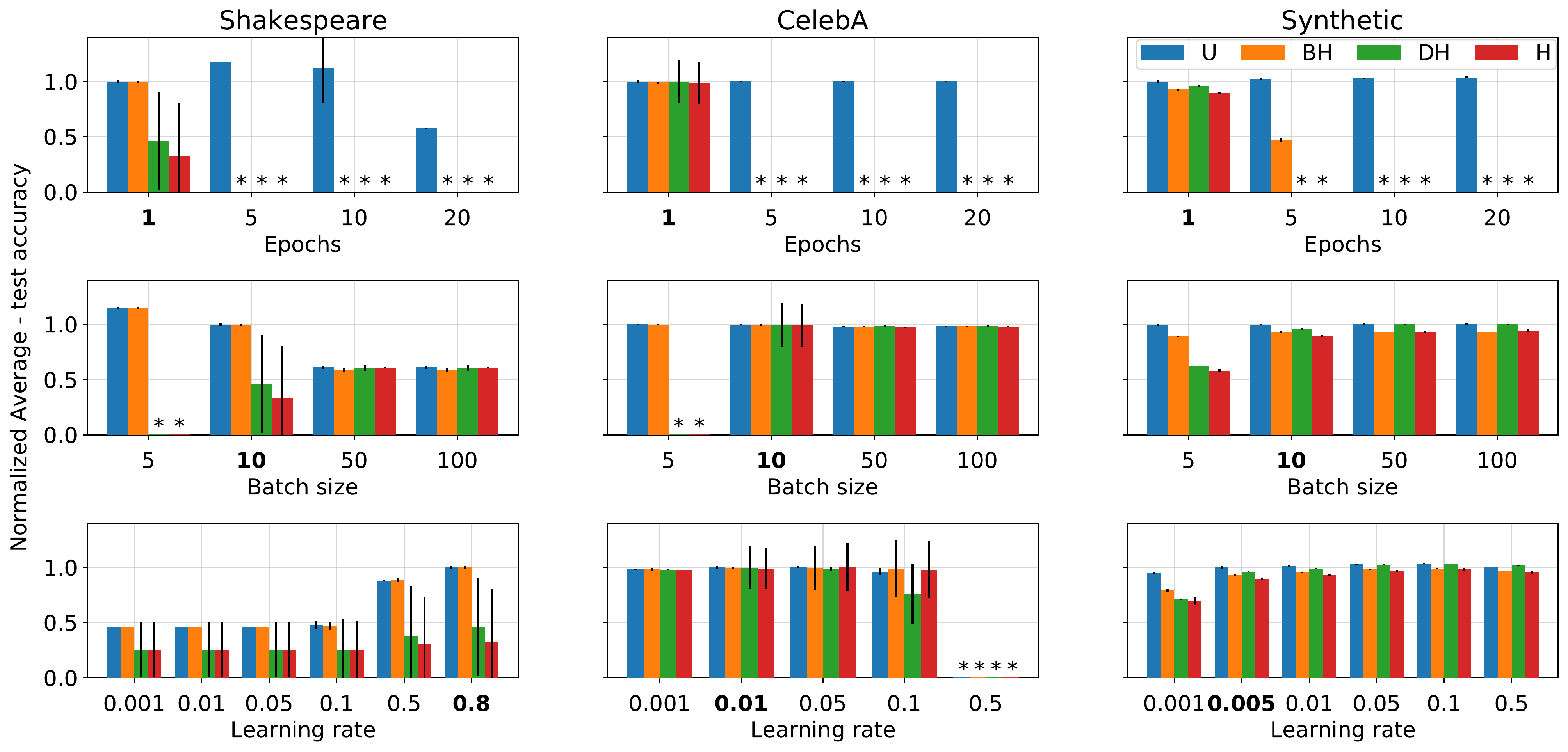}
    \caption{Shakespeare (Left), Celeb (Middle) and Synthetic (Right).}
    \end{subfigure}
    \caption{The sensitivity of normalized average test accuracy to the choice of the application-specific hyper-parameter for various benchmarks. Epochs (Top), Batch size (Middle), and Learning rate (Bottom) rows.}
    \label{fig:hyper-test-acc-train-params}
\end{figure}

\smartparagraph{Number of local epochs on the clients: } 
We observe that, for \UNI case, the average model accuracy and fairness nearly sees slight improvements with the increase of the number of epochs executed on clients' device (esp., Reddit and Shakespeare). However, for the heterogeneous settings (\BH, \DH, and \H), the performance of the model is extremely sensitive to the choice of the number of epochs. Specifically, the performance is significantly degraded leading to the divergence in nearly all heterogeneous cases. This is because of the additional computational time needed for clients to finish the extra epochs. This extra time renders many clients (mostly the low-end and moderate devices) unable to finish the training before the reporting deadline. Recall that, the server aggregates the updates to the global model if the number of received updates meets the target fraction. As a consequence, in many of the global rounds, the server abandons the round because the clients who successfully upload, in many of the global rounds, are less than the target update fraction. As a result, the training process sees almost no actual learning and hence the model diverges. Therefore, in heterogeneous settings, the careful choice of number of epochs is essential as it plays a trade-off between better model quality and model divergence (i.e., due to excessive round failures). 

\smartparagraph{Batch size:} We observe that, for \UNI and \BH settings, the model quality and fairness improves with the lower batch sizes used for training on the clients. This is because, in general, low batch sizes help with better generalization~\cite{masters2018revisiting}. However, with low batch size, clients incur higher execution time to finish the training epoch. For both \DH and \H settings, the higher execution cost leads to more clients failing to submit their updates within the deadline and as a result the model diverges. On the other hand, for all settings, increasing the batch size tend to reduce the average test accuracy and the fairness among the clients. This is because large batch sizes cause the over-fitting problem~\cite{masters2018revisiting}. These results suggest that choice of batch size not only influences the learning process but can influence the impact of heterogeneity.

\smartparagraph{Learning rate:} The results suggest that, for all heterogeneous settings, the average accuracy and fairness is quite sensitive to the choice of the learning rate used by the optimizer. In general, learning rate has to be tuned carefully for each learning task which is consistent with the literature. And, if fine tuned, it has nearly no influence on the level of impact on performance introduced by the heterogeneity. However, we observe that, for \DH and \H settings, the impact from unsuitable learning rate is amplified further (e.g., at lower learning rates such as $lr=0.001$). This might be because low learning rates results in slower learning for the clients. And, since \DH and \H result in lower numbers of successful clients, then it is expected that the magnitude of the aggregated model updates become quite insignificant when a small learning rate is used. In contrast, the magnitude of the update could have been amplified if a larger learning rate were used during the training.

\smartparagraph{Takeaways:}
The optimizer choice is mostly dependent on the nature of the learning problem, but task designers should spend more effort in experimenting with it in heterogeneous FL environments. The number of epochs should be tuned adaptively to optimize the trade-off between any gains in performance and the impact from heterogeneity in the network. The choice of batch size has to be moderate as there is a trade-off between model performance (e.g., generalization) and execution time. In general, the batch size should be tuned along with the deadline. The deadline should be set large enough to allow the majority of the clients to finish in time the training with the chosen batch size. Lastly, the learning rate should be dynamically tuned for each round based on historical data on the expected success rate for different periods of the day.

\subsection{Effectiveness of FL algorithm variants}

\begin{table}[!h]
\small
\caption{Aggregation algorithms and their hyper-parameters.}
\centering
\resizebox{0.45\columnwidth}{!}{\begin{tabular}{ |c|c| } 
        \hline
        \textbf{Hyper-parameter} & \textbf{Values} \\
        \hline
        FL Algorithm & [FedAvg, FedProx, Q-FFL] \\ 
        \hline
        FedProx ($\mu$) & [0, 0.001, 0.01, 0.1, 0.5, 1] \\
        \hline
        Q-FFL ($q$) & [0, 0.001, 0.1, 1, 5] \\
        \hline
        \end{tabular}
}
\label{tab:agg-hyperparamters}
\end{table}

\begin{figure}[!t]
  \centering
    \begin{subfigure}{0.78\linewidth}
    \includegraphics[width=1\linewidth]{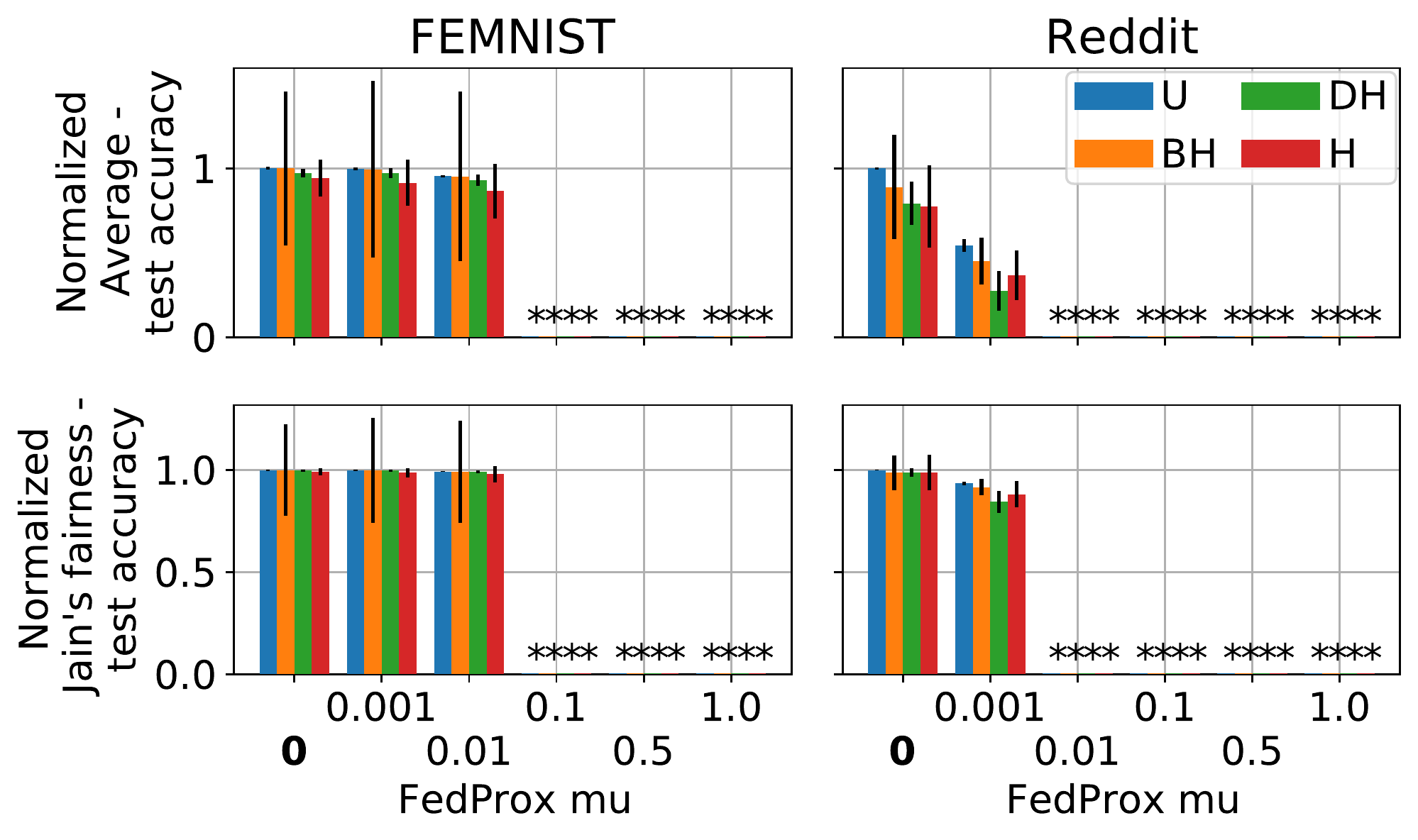}
    \caption{FEMNIST (Left) and Reddit (Right). }
    \end{subfigure}
    \begin{subfigure}{1\linewidth}
     \includegraphics[width=1\linewidth]{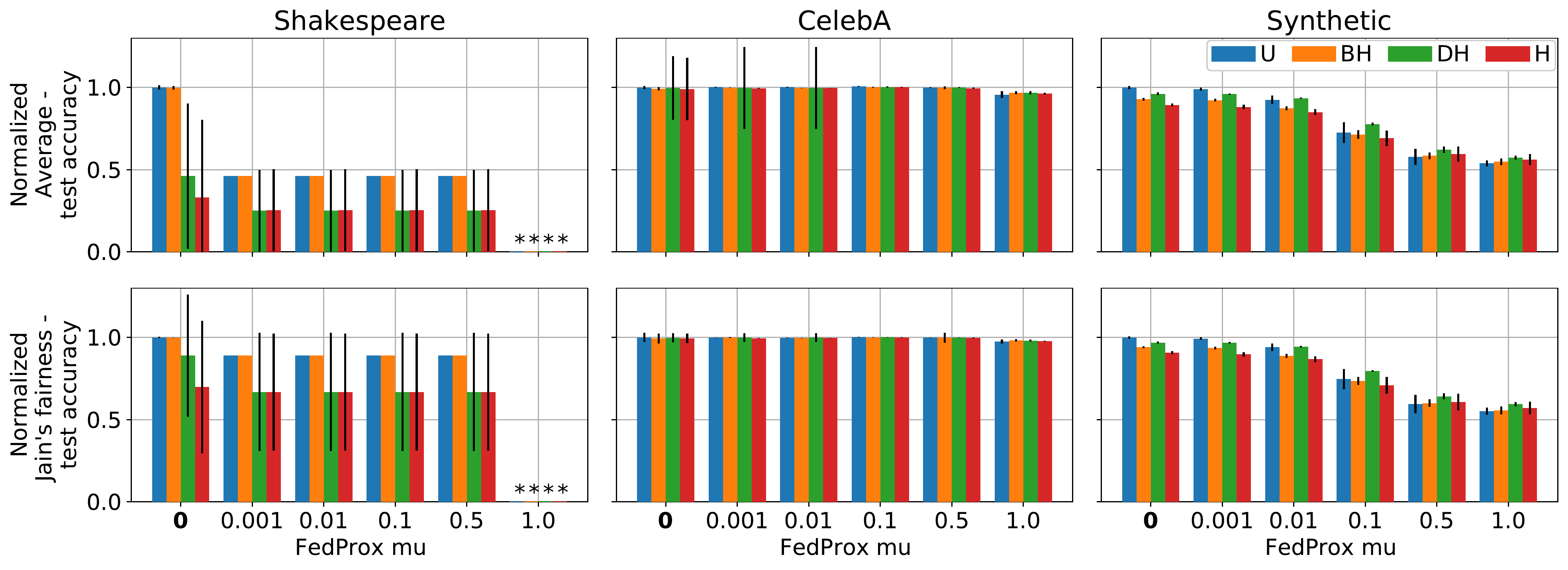}
     \caption{Shakespeare (Left), Celeb (Middle) and Synthetic (Right).}
    \end{subfigure}
    \caption{Performance of FedProx algorithm. The normalized average test accuracy (Top Rows) and normalized Jain's Fairness (Bottom Rows)}
    \label{fig:fedprox-heter-impact}
\end{figure}

We compare FedProx~\cite{Li2020FedProx} and Q-FFL~\cite{Li2020Fair} to the baseline FedAvg algorithm~(see \cref{apdx:algo} for pseudo-codes). We explore the hyper-parameter values that were used in these works (see \cref{tab:agg-hyperparamters}). 

\smartparagraph{FedProx Algorithm:} We observe in \cref{fig:fedprox-heter-impact}, for all settings, that FedProx algorithm does not improve both the average accuracy and fairness compared to FedAvg. More importantly, FedProx leads to degradation in model performance and model divergences for certain $\mu$ settings. These observations are consistent for all the benchmarks except for CelebA where FedProx results in neither any improvement nor noticeable degradation. Even though the proximal term $\mu$ is designed to absorb the impact from heterogeneous local updates, the aforementioned observations are consistent for all benchmarks in our study and for various settings of the hyper-parameter of FedProx algorithm (i.e., the proximal term $\mu$). 

\begin{figure}[!t]
  \centering
    \begin{subfigure}{0.78\linewidth}
    \includegraphics[width=1\linewidth]{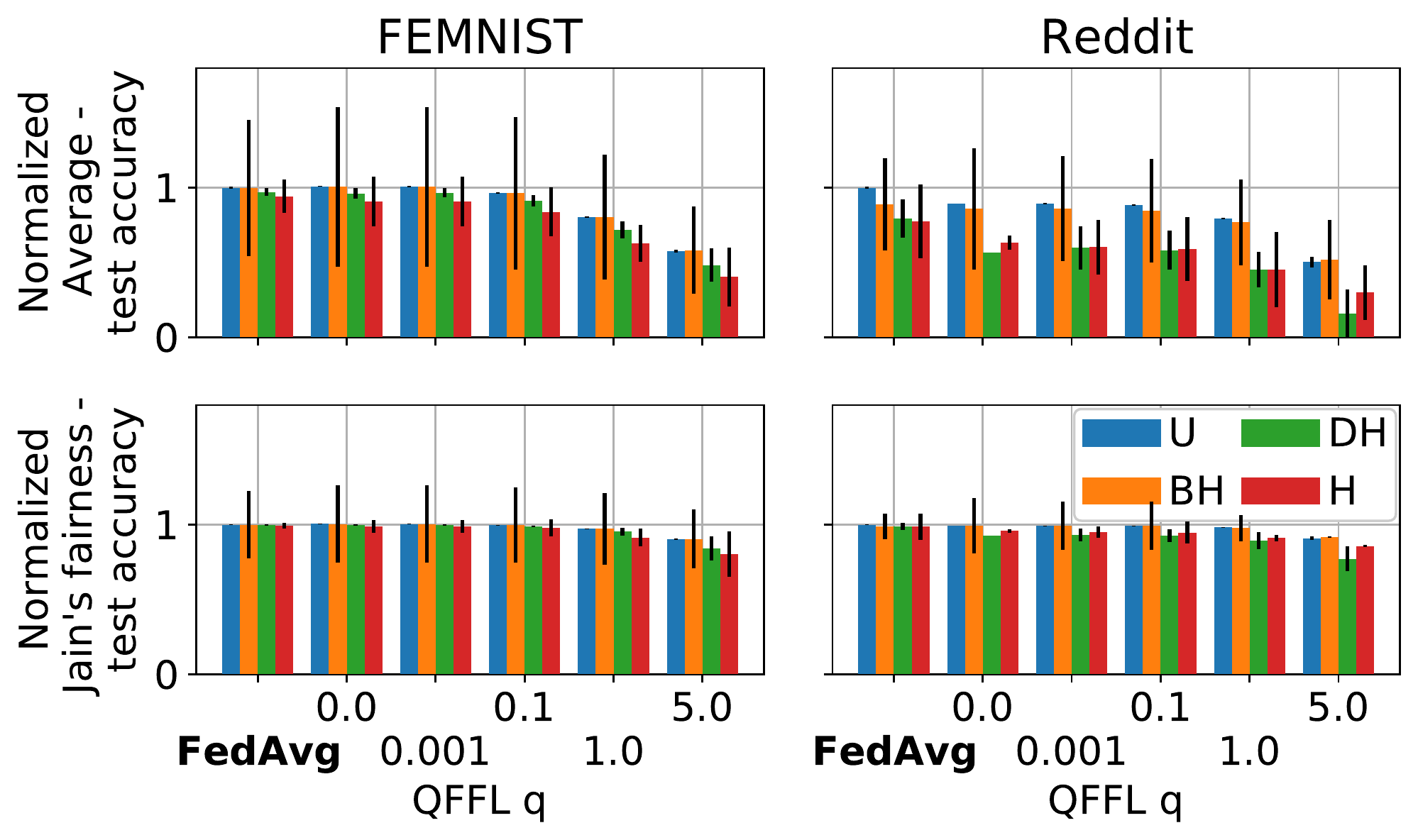}
    \caption{FEMNIST (Left) and Reddit (Right). }
    \end{subfigure}
    \\
    \begin{subfigure}{1\linewidth}
    \includegraphics[width=1\linewidth]{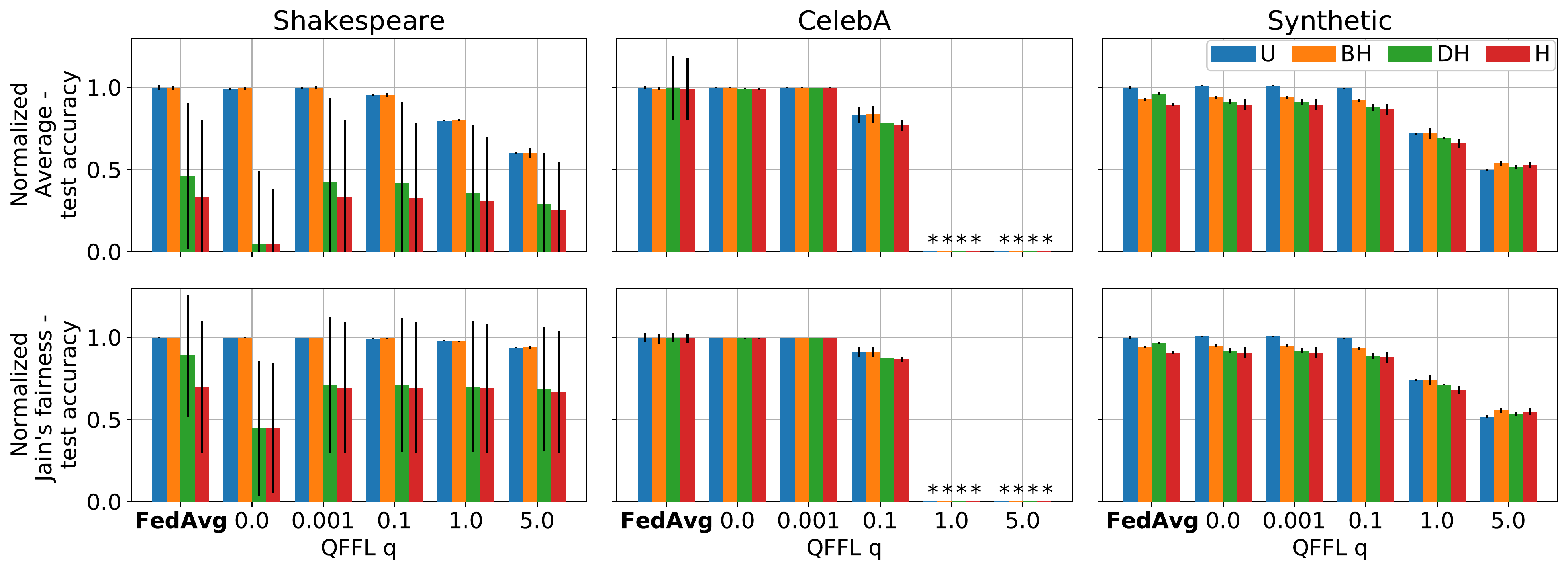}
    \caption{Shakespeare (Left), Celeb (Middle) and Synthetic (Right).}
    \end{subfigure}
    \caption{Performance of Q-FFL algorithm. The normalized average test accuracy (Top Rows) and normalized Jain's Fairness (Bottom Rows).}
    \label{fig:qffl-heter-impact}
\end{figure}

\smartparagraph{Q-FFL Algorithm:} \cref{fig:qffl-heter-impact} shows that the average accuracy and fairness of the global model are not improved as a result of employing the Q-FFL algorithm in all settings compared to FedAvg. Worse, the performance and fairness, in most cases, are severely degraded. We observe that for small values of $q$, neither the average nor the fairness is improved. However, as $q$ increases, both performance measures are decreased significantly.\footnote{AFL~\cite{Mohri2019AgnosticFL} is obtained by setting large $q$ value for Q-FFL; hence the observations for $q$ value of 5 applies to AFL.}  Notably, for \DH and \H settings, the impact is further amplified. Finally, these results are consistent over all the benchmarks. We note that theses observations also align with the remarks made for FedProx and Q-FFL in ~\cite{yang2020heterogeneityaware}. 

\smartparagraph{Takeaways:} We attribute these results, however, to the differences in the evaluation scenarios and settings in both FedProx and Q-FFL. Moreover, device and behavioral heterogeneity is influenced mostly by system than algorithmic hyper-parameters, hence algorithmic solutions may not be quite effective. However, these observations encourage further exploration on the direction of devising adaptive system designs to tackle the inherent challenges and performance impacts as a result of the clients' behavior, hardware and network heterogeneity in the FL environments.

\section{Related Work}
\label{sec:related}

\smartparagraph{Federated Learning (FL):} is a paradigm that has recently emerged mainly for privacy-preserving learning. In this paradigm, training is distributed on decentralized devices such as smart edge devices (i.e., mobile or IoT sensor devices) which produces data samples, so that data does not leave the local storage of the data source~\cite{li2014scaling,mcmahan2017}. For this reason, FL has seen an emerging popularity and is currently deployed for a large number of users to enhance the functionality of the virtual keyboards (e.g., the search suggestion quality\cite{Bonawitz19,yang2018applied}). Moreover, to facilitate and expedite research efforts, several works developed FL frameworks for experimentation with FL settings \cite{caldas2018leaf,ryffel2018generic,paddle,tff,yang2020heterogeneityaware}. 
Flash~\cite{yang2020heterogeneityaware} is a recently developed platform based on Leaf~\cite{caldas2018leaf}, which incorporates heterogeneity-related parameters into the design. In this work, we dissect heterogeneity and provide a comprehensive evaluation of their impact on the model quality and fairness.

\smartparagraph{System heterogeneity:} Heterogeneity is one of the major challenges for distributed systems.
In datacenters, the compute nodes need to aggregate their local model updates among each other via some form of communication back-end (e.g., using parameter server \cite{li2014scaling,bytescheduler} or peer-to-peer collective aggregation \cite{horovod,Patarasuk2016,plink-mlsys2020}). In this context, device heterogeneity results in performance degradation due to stragglers (i.e., slow workers) who slow down the training process~\cite{Jiang2017,Chen2018}. Several works tried to address this problem via system and algorithmic solutions~\cite{hogwild,Qirong2013,Jianmin2016,Patarasuk2016,Jiang2017,Chen2018}. In FL settings, the heterogeneity is sourced from other system artifacts and is not limited to the heterogeneity in device capabilities. Specifically, data distribution among the clients, client sampling method, and user behavior are other main sources of heterogeneity in FL scenarios. This work focuses on heterogeneity in FL settings and evaluates its impact on the learning process and the trained model. Our evaluation shows that device and behavioral heterogeneity can have severe negative impact on model performance, such as divergence in the worst cases.

\smartparagraph{Improvements in FL:} 
In FL, some proposals try to address the communication bottlenecks during the training via exploiting number of communication reduction techniques like compression, periodic update, and layer-wise asynchronous updates
\cite{Jakub2016,Smith2017,Bonawitz19,Chen2019CommunicationEfficientFD,pmlr-v108-reisizadeh20a}. Other works try to study and improve the privacy guarantees of FL environments \cite{McMahan2018,Melis2019,Bonawitz19,Nasr2019,Bagdasaryan20}. Additionally, others focus on personalizing the global model resulting from FL~\cite{jiang2019improving} and minimizing the energy consumption on edge devices~\cite{Li2019smartpc}. Moreover, recent works highlighted the problems of bias and proposed mitigation schemes to enforce fairer representation of the clients in the trained model~\cite{Mohri2019AgnosticFL,Li2020Fair,Li2020FedProx}. For instance, the authors in \cite{Mohri2019AgnosticFL} optimize the central model for any target data distribution formed by the mixture of the clients' data distributions. However, our evaluation shows that these algorithms can not effectively mitigate the impact of device and behavioral heterogeneity. In this work, our empirical evaluation of various heterogeneous settings show that the existing proposals fall short in mitigating the impact of device and behavioral heterogeneity on the performance and fairness of the trained models.

\section{Conclusion}
\label{sec:conclusion}

We present an extensive experimental study of the potential impact of heterogeneous participants on the performance and level of bias of the collaboratively learnt models in federated learning settings. To this end, we empirically study the factors that play a main role in introducing heterogeneity, such as device and behavioral heterogeneity. We evaluate how different hyper-parameters of the FL process amplify the impact of heterogeneity on both model performance and fairness. Our evaluation, which spans a wide range of configurations and benchmarks, shows that heterogeneity can cause up to 4.6$\times$ and 2.2$\times$ degradation in the average test accuracy and the fairness of the model trained in heterogeneous (\H) setting compared to one trained in uniform (\UNI) setting. We envision that our work will benefit:
\begin{inparaenum}[(1)]
    \item Researchers, who will use our study as the basis for more efficient algorithms that are amenable to heterogeneity;
    \item and Practitioners, who need to tune their FL applications deployed in real-world environments to achieve better fairness and performance.
\end{inparaenum}

\bibliography{references}
\bibliographystyle{abbrv}

\clearpage
\appendix

\section{Datasets and Models}
\label{apdx:datasets}
Here we provide full details on the datasets and models used in our experiments. We curate a diverse set of non-synthetic datasets, including those used in prior work on federated learning~\cite{mcmahan2017}, and some proposed in LEAF, a benchmark for FL settings~\cite{caldas2018leaf}. In Table \ref{tab:data-stats}, we summarize the statistics of our data, as they have been formulated after the processing they have undergone.

\smartparagraph{Synthetic:} We generate two IID and non-IID datasets similar to~\cite{caldas2018leaf}. For all synthetic datasets, the data is classified into 10 classes. A total of 10,000 clients are used and the number of samples on each client follows a power law.

\smartparagraph{FEMNIST:} We study an image classification problem on the 62-class EMNIST dataset~\cite{EMNIST} using multinomial logistic regression. To generate heterogeneous data partitions, we subsample 10 lower case characters (a-j) from EMNIST and distribute only 5 classes to each device. We call this federated version of EMNIST as FEMNIST. There are 200 devices in total. The input of the model is a flattened 784-dimensional $(28 \times 28)$ image, and the output is a class label between 0 and 9.

\smartparagraph{CelebA:} This binary image classification task determines whether a celebrity in the image is smiling or not. The images are sourced from~\cite{CELEBA}. To ensure good convergence, celebrities with less than 5 images in the dataset are ignored during the pre-processing stage.

\smartparagraph{Shakespeare:} This is a dataset built from The Complete Works of William Shakespeare~\cite{mcmahan2017}. Each speaking role in a play represents a different device. We use a two-layer LSTM classifier containing 100 hidden units with an 8D embedding layer. The task is next-character prediction, and there are 80 classes of characters in total. The model takes as input a sequence of 80 characters, embeds each of the characters into a learned 8-dimensional space and outputs one character per training sample after 2 LSTM layers and a densely-connected layer.

\smartparagraph{Reddit:} This is a dataset built from the answers of the users on the reddit threads. The Reddit data which is released by~\cite{REDDIT} corresponds to the collected data on December 2017. To preprocess the dataset, the following operations are preformed:
\begin{inparaenum}[1)]
\item Unescape HTML symbols;
\item Remove extraneous whitespaces;
\item Remove non-ASCII symbols;
\item Replace URLs, reddit usernames and subreddit names with special tokens;
\item Lowercase the text;
\item Tokenize the text (using nltk's TweetTokenizer).
\end{inparaenum}
We also remove users and comments that simple heuristics or preliminary inspections mark as bots. Due to the size of the dataset, users with less than 5 or more than 1000 comments are removed (which account for less than $0.01\%$ of users). Finally, The data is further processed to make it ready for the reference model (i.e., by splitting it into train/validation/test sets and by creating sequences of 10 tokens for the LSTM model in use)~\cite{yang2020heterogeneityaware}. A vocabulary of the 10 thousand most common tokens in the data is also used for labelling purposes.

\begin{table*}[!t]
\small
\centering
\caption{Dataset Statistics}
\resizebox{0.9\linewidth}{!}{\begin{tabular}{ *8c } 
        \toprule
         \textbf{Dataset} & \textbf{Clients} & \makecell{\textbf{Data Split}  \\ \textbf{Train\%-Test\%}} & \makecell{\textbf{Total} \\ \textbf{Samples}} & \makecell{\textbf{Samples}/\ \\ \textbf{Users}}	& \multicolumn{3}{c}{\textbf{\# of Samples}} \\
          & & & & & std & std /\ mean & skewness \\
         \hline
          FEMNIST & 3,400 & 90\%-10\% & 748,698 & 220.21 & 85.04 & 0.39 & 0.88 \\
         \hline
         CelebA & 9,343 & 90\%-10\% & 200,288 & 21.44 & 7.63 & 0.36 & -0.54 \\
         \hline 
         Shakespeare & 1,129 & 80\%-20\%  & 4,226,158 & 3,743.28 & 6212.26 & 1.66 & 3.35 \\
         \hline
         Reddit & 813 & 50\%-50\% & 16,362 & 20.03 & 36.07 & 1.80 & 5.06 \\
         \hline
         Synthetic & 10,000 & 90\%-10\% & 1,011,245 & 101.12 & 201.03 & 1.99 & 3.26 \\
        \bottomrule
        \end{tabular}
}
\label{tab:data-stats}
\end{table*}

\smartparagraph{Data Sampling}
For each dataset, we introduce two sample methods, IID and non-IID:
\begin{itemize}
    \item In the IID sampling scenario, each data point is equally likely to be sampled. Thus, all users have the same underlying distribution of data.
    \item In the non-IID sampling scenario, the underlying distribution of data for each user is consistent with the raw data.
\end{itemize}

\subsection{Data Distribution}
\begin{figure*}[t]
  \centering
    \begin{subfigure}[ht]{0.32\textwidth}
  \includegraphics[width=1\textwidth]{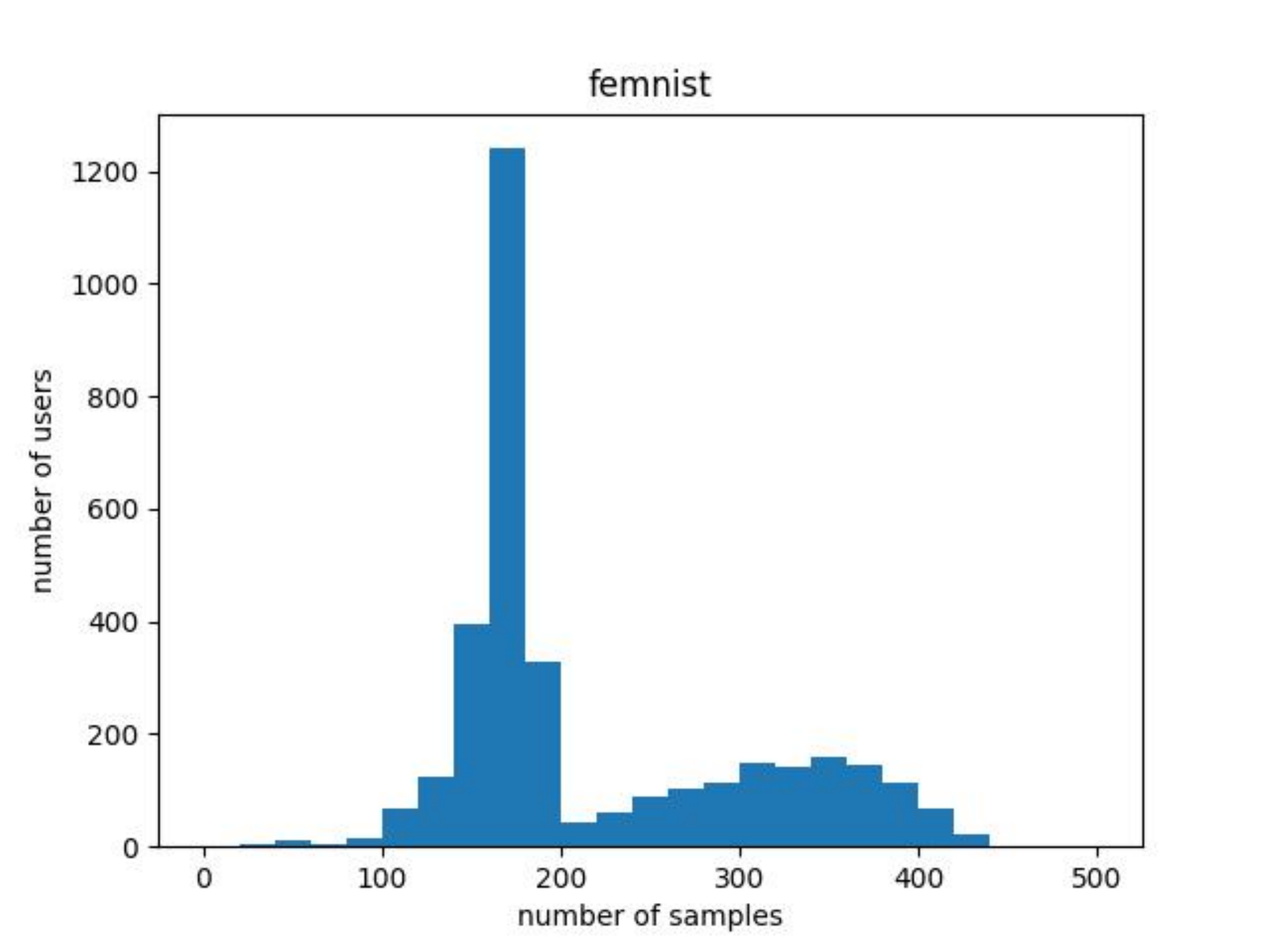}
  \caption{Femnist}
    \label{fig:femnist-hist}
     \end{subfigure}
    \begin{subfigure}[ht]{0.32\textwidth}
  \includegraphics[width=1\textwidth]{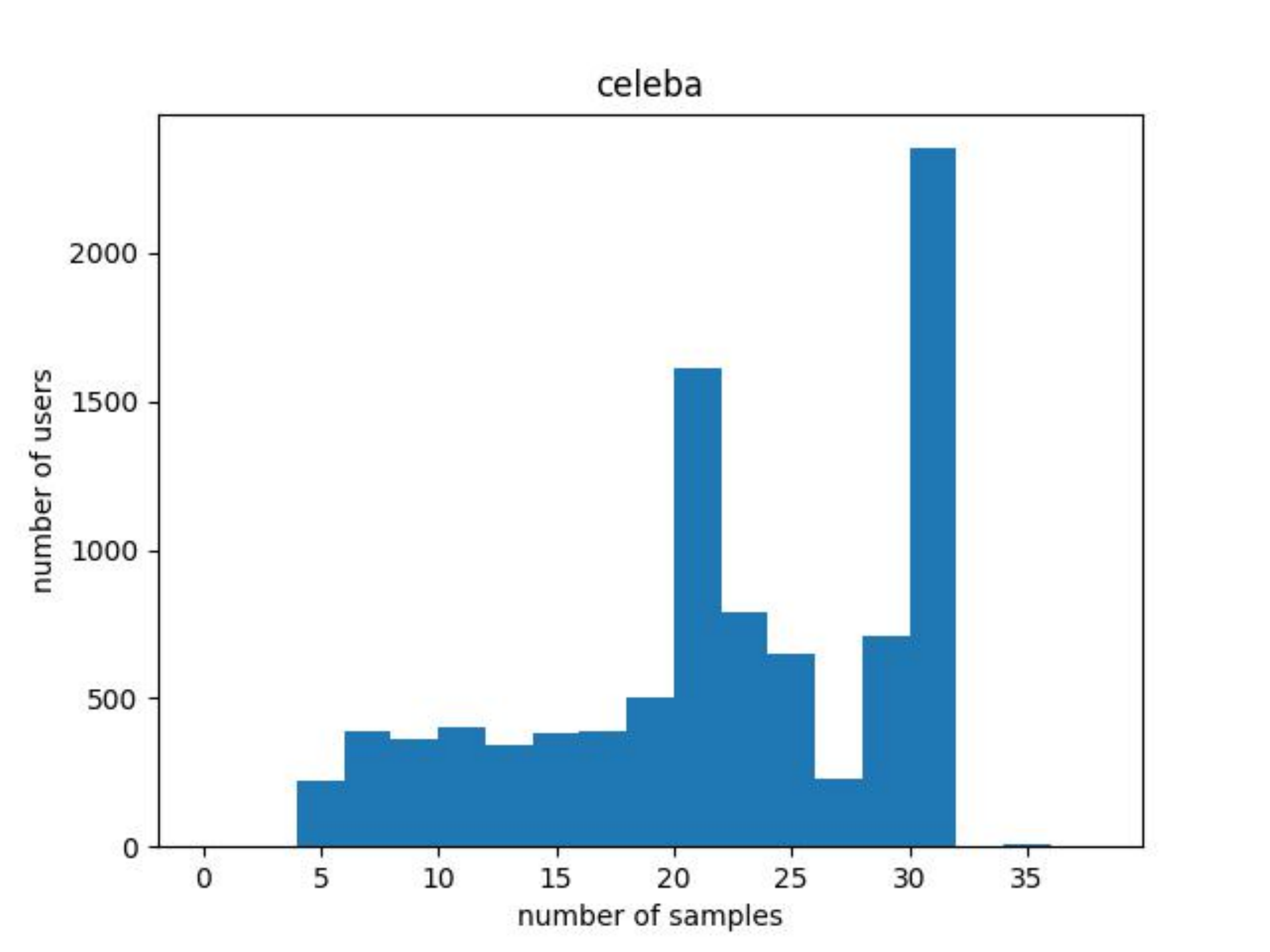}
    \caption{Celeba}
     \label{fig:celeba-hist}
    \end{subfigure}
        \\
  \begin{subfigure}[ht]{0.32\linewidth}
     \includegraphics[width=1\textwidth]{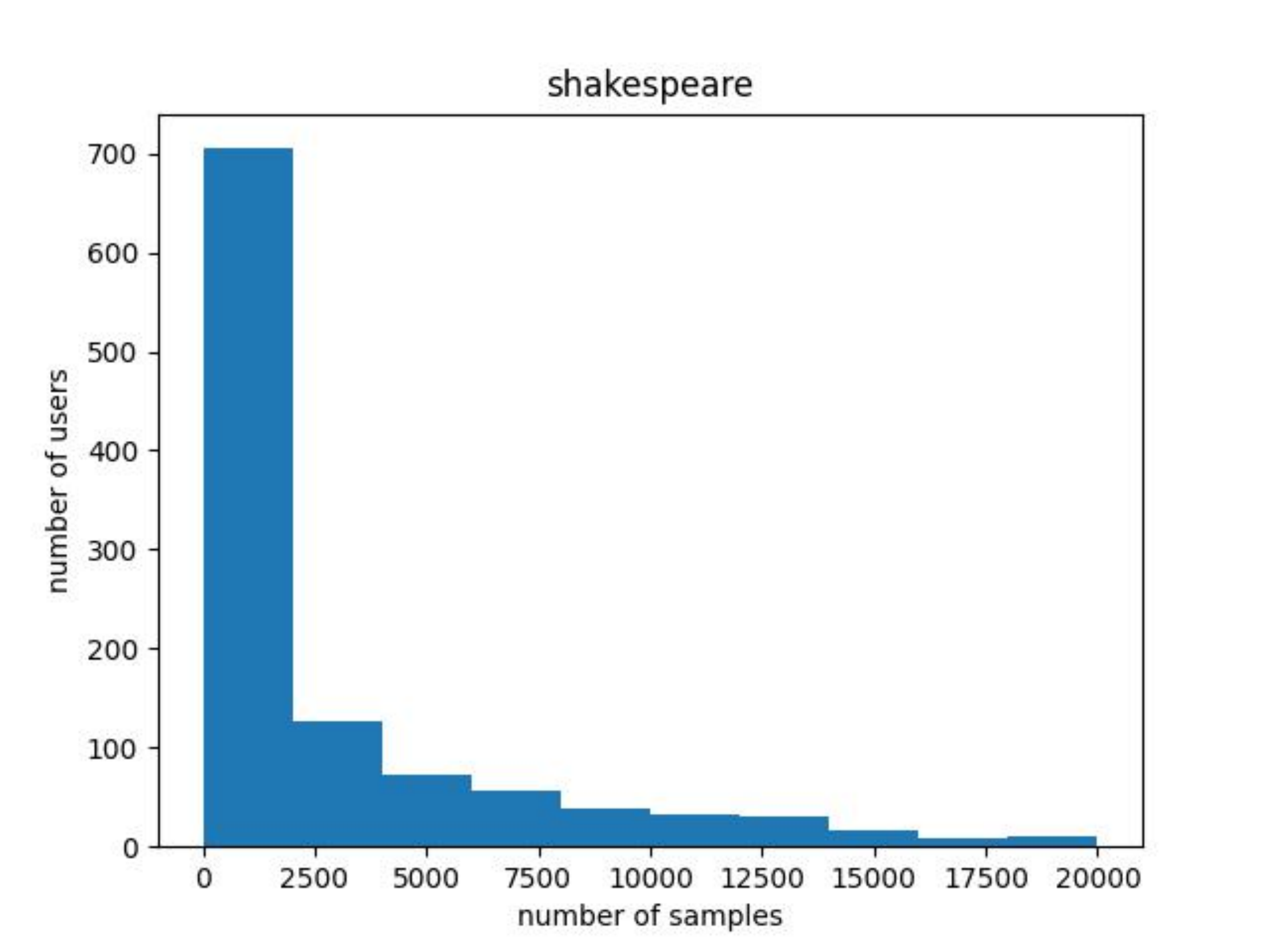}
    	\caption{Shakespeare}
     \label{fig:shakespeare-hist}
    \end{subfigure}
    \begin{subfigure}[ht]{0.32\linewidth}
  \includegraphics[width=1\textwidth]{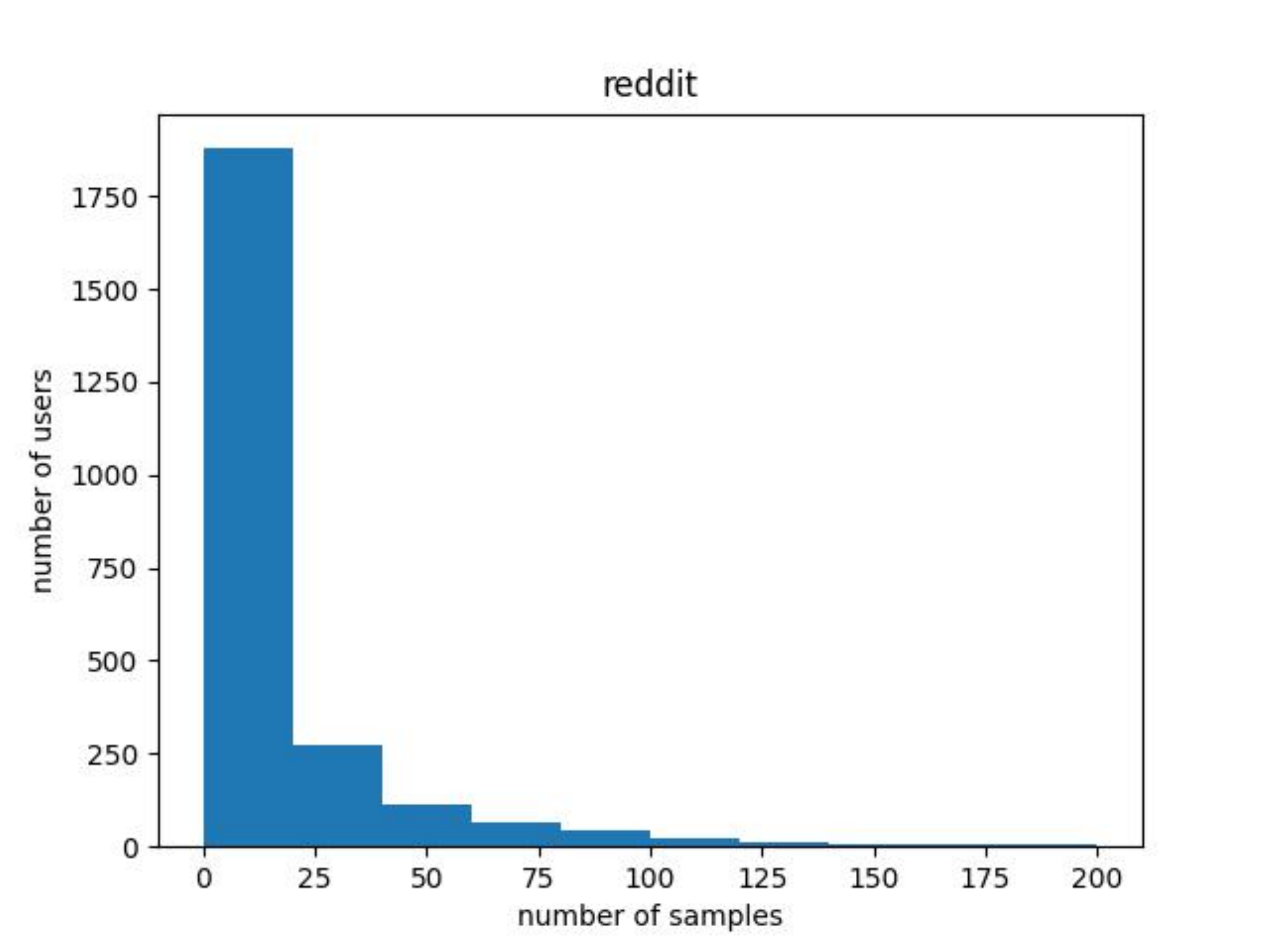}
    \caption{Reddit}
     \label{fig:reddit-hist}
    \end{subfigure}
    \hfill
     \begin{subfigure}[ht]{0.32\linewidth}
  \includegraphics[width=1\textwidth]{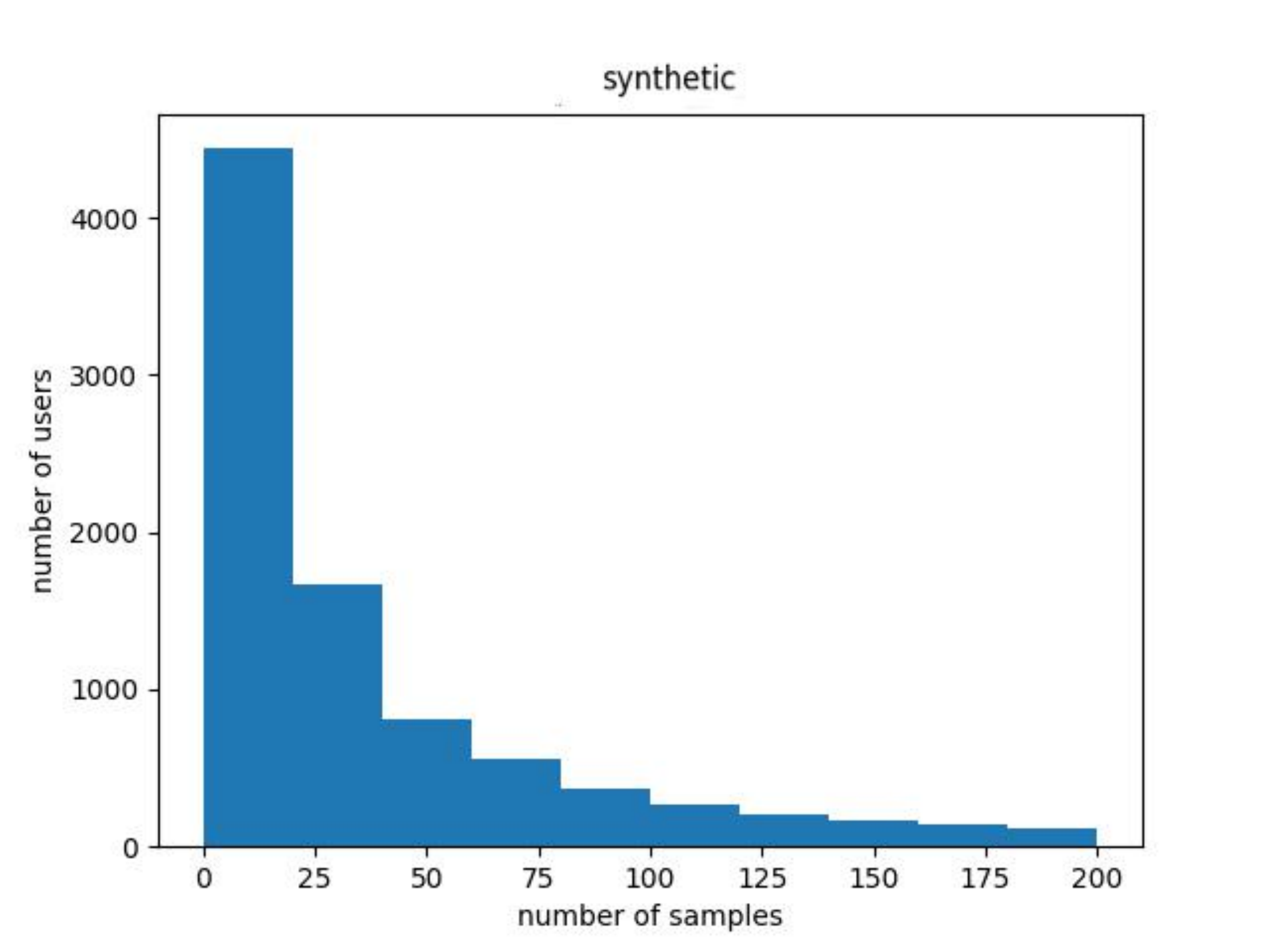}
    \caption{Synthetic}
     \label{fig:synthetic-hist}
    \end{subfigure}
    \caption{Data samples distribution for FL benchmarks }
     \label{fig:data-distribution}
\end{figure*}

Next, we provide plots for each dataset that demonstrate the distribution of data samples among the total number of users.
As can be clearly seen in Figure \ref{fig:data-distribution}, both the FEMNIST and CelebA datasets do not follow a normal distribution. In contrast, the Shakespeare, Reddit and Synthetic datasets seem to have a long tailed distribution with most of the users allocated similar number of samples. 

\section{Experimental Environment}
\label{apdx:clusters}

We run the large number of experiments leveraging GPUs available within two clusters. One cluster is under our control (i.e., Dedicated) and another is allocation-based (i.e., Shared). We allocate one GPU to run each experiment. The system specification of the servers is described below:

\paragraph{\textbf{Cluster 1 - Dedicated Environment}}
\begin{itemize}
    \item Tesla P100 or V100-SXM2 with 16GB of GPU memory
    \item CPU: Intel(R) Xeon(R) Silver 4112 CPU @ 2.60GHz, 16 cores
    \item System memory: 512 GiB
    \item OS: Ubuntu 18.04 + Linux Kernel v4.15
    \item Environment: Miniconda 4.3, CUDA 10.1, CUDNN 7.6 
    \item Software: Python 3.6, TensorFlow 1.14.0, Wandb 0.8.35
    
\end{itemize}
\paragraph{\textbf{Cluster 2 - Shared Environment}}
\begin{itemize}
    \item Nvidia Tesla P100 with 16GB or V100-SXM2 with 32GB of GPU memory
    \item CPU:  Intel(R) Xeon(R) Gold 6248 CPU @ 2.50GHz, 16 cores
    \item System memory: 512 GiB
    \item OS: CentOS 7.7 + Linux Kernel v3.10
    \item Environment: Miniconda 4.3, CUDA 10.1, CUDNN 7.6 
    \item Software: Python 3.6, TensorFlow 1.14.0, Wandb 0.8.35
\end{itemize}

\section{Pseudo-code of FL Algorithms}
\label{apdx:algo}

\begin{algorithm*}[!h]
\caption{FedAvg~\cite{mcmahan2017}}\label{alg:fedavg}
\begin{algorithmic}
  \State \textbf{Input:} $K, T, \eta, E, w^0, N, p_k, k = 1, \dots, N$
  \For {$t = 0, \dots, T - 1$}
  \State Server selects at random a subset $S_t$ of $K$ devices (each device is selected with probability $p_k$)
  \State Server sends $w^t$ to all selected devices
  \State Each selected device, $k \in S_t$, runs SGD for $E$ epochs on $F_k$ with step-size $\eta$ to obtain $w^t$
  \State Each selected device $k$ sends $w_{k}^{t+1}$ back to the server
  \State Server aggregates the $w$'s as $w^{t+1} = \frac{1}{K} \sum \mathop{}_{\mkern-5mu k \in S_t} {w_{k}^{t+1}} $
  \EndFor
\end{algorithmic}
\end{algorithm*}

\begin{algorithm*}[!h]
\caption{FedProx~\cite{Li2020FedProx}}\label{alg:fedprox}
\begin{algorithmic}
  \State \textbf{Input:} $K, T, \mu, \gamma, w^0, N, p_k, k = 1, \dots, N$
  \For {$t = 0, \dots, T - 1$}
  \State Server selects at random a subset $S_t$ of $K$ devices (each device is selected with probability $p_k$)
  \State Server sends $w^t$ to all selected devices
  \State Each selected device, $k \in S_t$, finds a $w_{k}^{t+1}$ which is $\gamma_{k}^t$ - inexact minimizer of:
  \State \hspace*{5mm} $w_{k}^{t+1} \approx \arg \min_{w} h_{k}(w; w^t) = F_k(w) + \frac{\mu}{2} \| w - w^t \|^{2}$ 
  \State Each selected device $k$ sends $w_{k}^{t+1}$ back to the server
  \State Server aggregates the $w$'s as $w^{t+1} = \frac{1}{K} \sum \mathop{}_{\mkern-5mu k \in S_t} {w_{k}^{t+1}} $
  \EndFor
\end{algorithmic}
\end{algorithm*}

\begin{algorithm*}[!h]
\caption{q-FedAvg~\cite{Li2020Fair}}\label{alg:q-fedavg}
\begin{algorithmic}
  \State \textbf{Input:} $K, T, \eta, E, q, \frac{1}{L}, w^0, p_k, k = 1, \dots, m$
  \For {$t = 0, \dots, T - 1$}
  \State Server selects at random a subset $S_t$ of $K$ devices (each device is selected with probability $p_k$)
  \State Server sends $w^t$ to all selected devices
  \State Each selected device, $k \in S_t$, runs SGD for $E$ epochs on $F_k$ with step-size $\eta$ to obtain $w^t$
  \State Each selected device $k \in S_t$ computes: 
  \State \hspace*{5mm} $\Delta w_{k}^t = L(w^t - \bar{w}_{k}^{t+1})$
  \State \hspace*{5mm} $\Delta_{k}^t = F_{k}^q(w^t)\Delta w_{k}^t$
  \State \hspace*{5mm} $h_{k}^t = q F_{k}^{q-1}(w^t) \| \Delta{w_{k}^t} \|^{2} + L F_{k}^q(w^t)$ 
  \State Each selected device $k$ sends $\Delta_{k}^t$ and $h_{k}^t$ back to the server
  \State Server aggregates the $w$'s as $w^{t+1} = w^{t} - \frac{\sum \mathop{}_{\mkern-5mu k \in S_t} {\Delta_{k}^t }} {\sum \mathop{}_{\mkern-5mu k \in S_t} {h_{k}^t}}$
  \EndFor
\end{algorithmic}
\end{algorithm*}

\begin{figure}[t]
  \centering
    \includegraphics[width=1\linewidth]{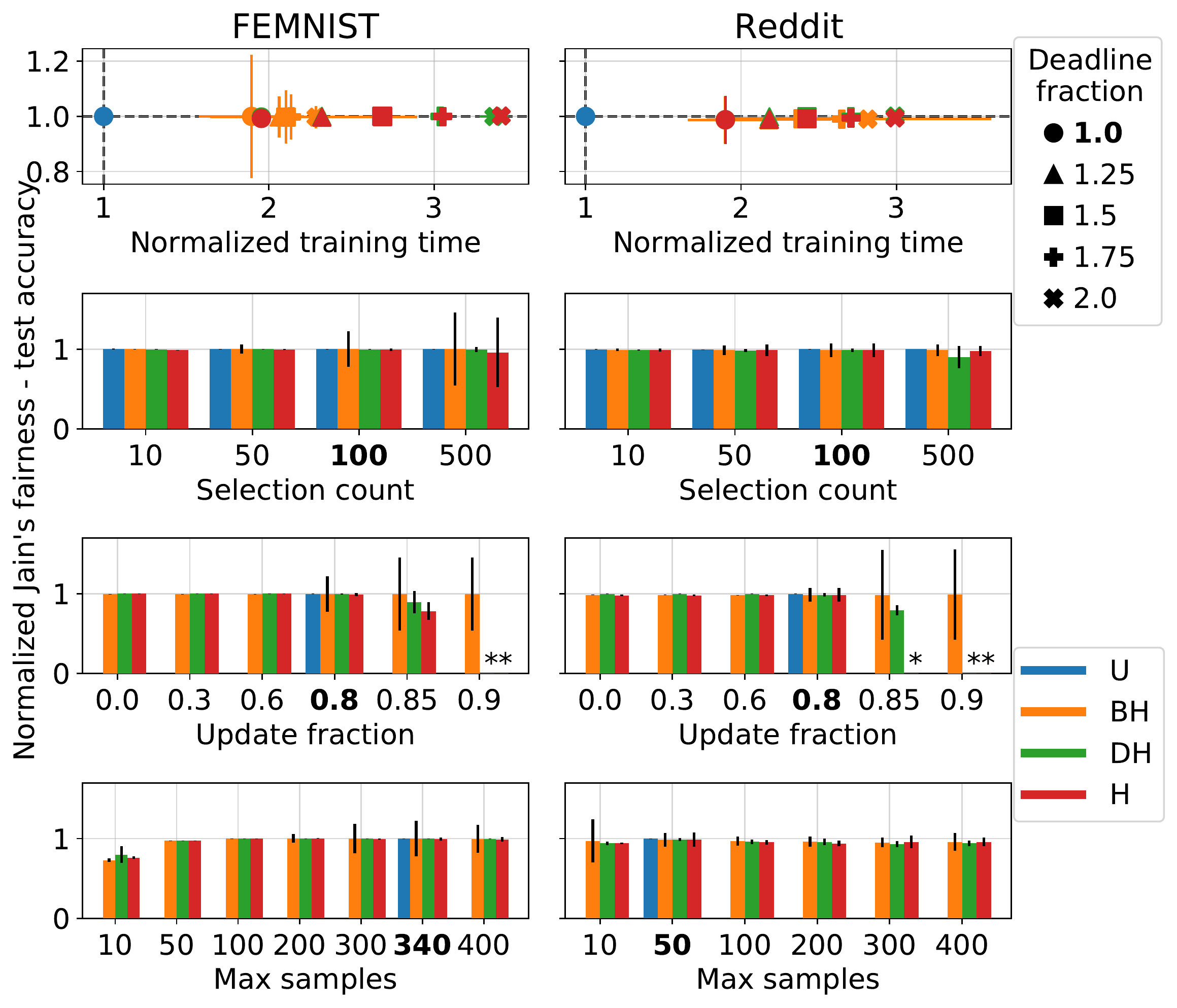}
    \caption{The sensitivity of the model fairness to the choice of the FL hyper-parameters for FEMNIST and Reddit Benchmarks. We mark the default settings of the uniform scenario in bold in all following figures.}
    \label{fig:fairness-test-acc-FL-params}
\end{figure}

\begin{figure*}[!ht]
  \centering
    \includegraphics[width=1\linewidth]{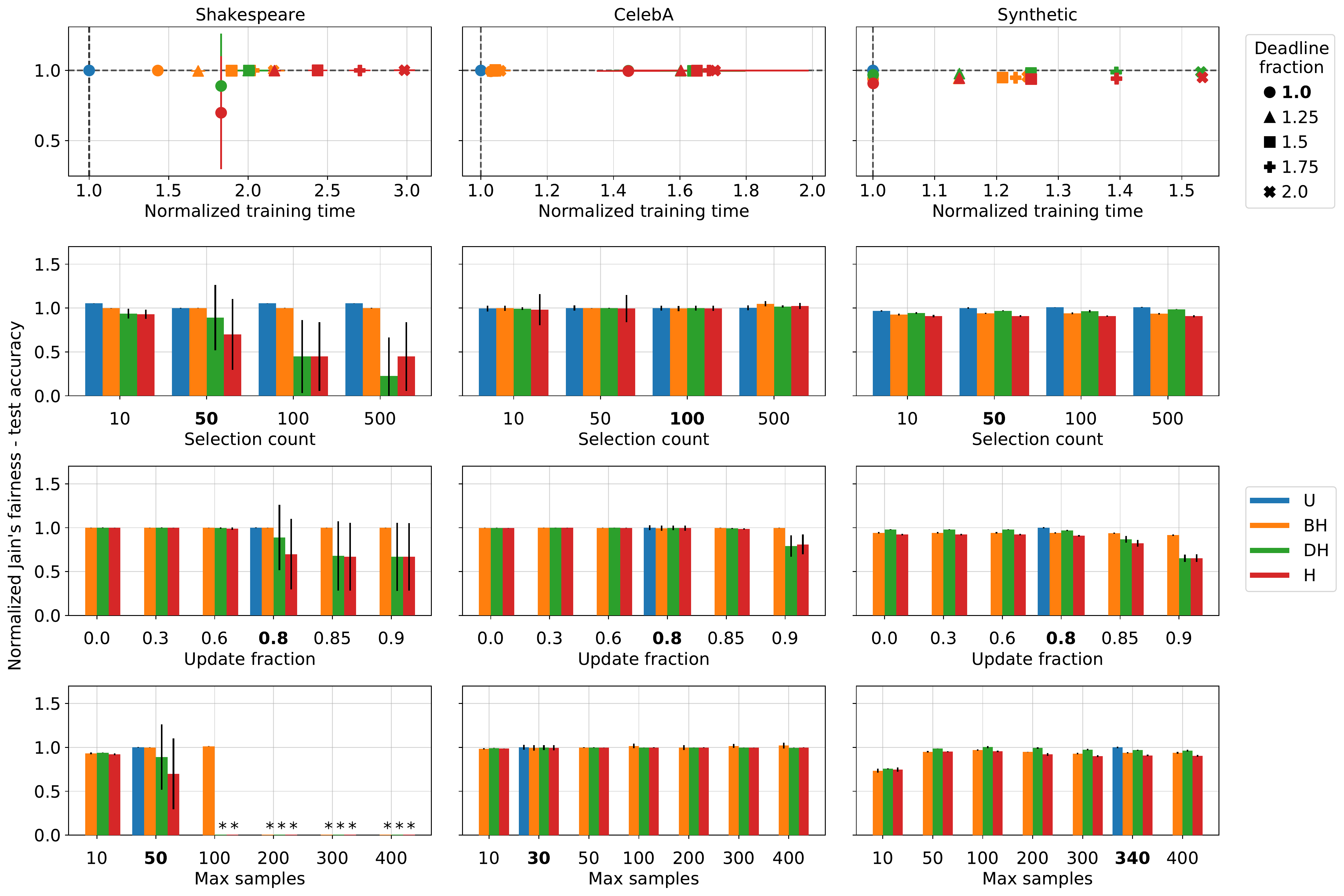}
    \caption{Sensitivity of the model fairness to the choice of the FL hyper-parameters for Shakespeare, CelebA and Synthetic Benchmarks.}
    \label{fig:app:fairness-test-acc-FL-params-apdx}
\end{figure*}

\begin{figure}[!t]
  \centering
  \begin{subfigure}{0.78\linewidth}
   \includegraphics[width=1\linewidth]{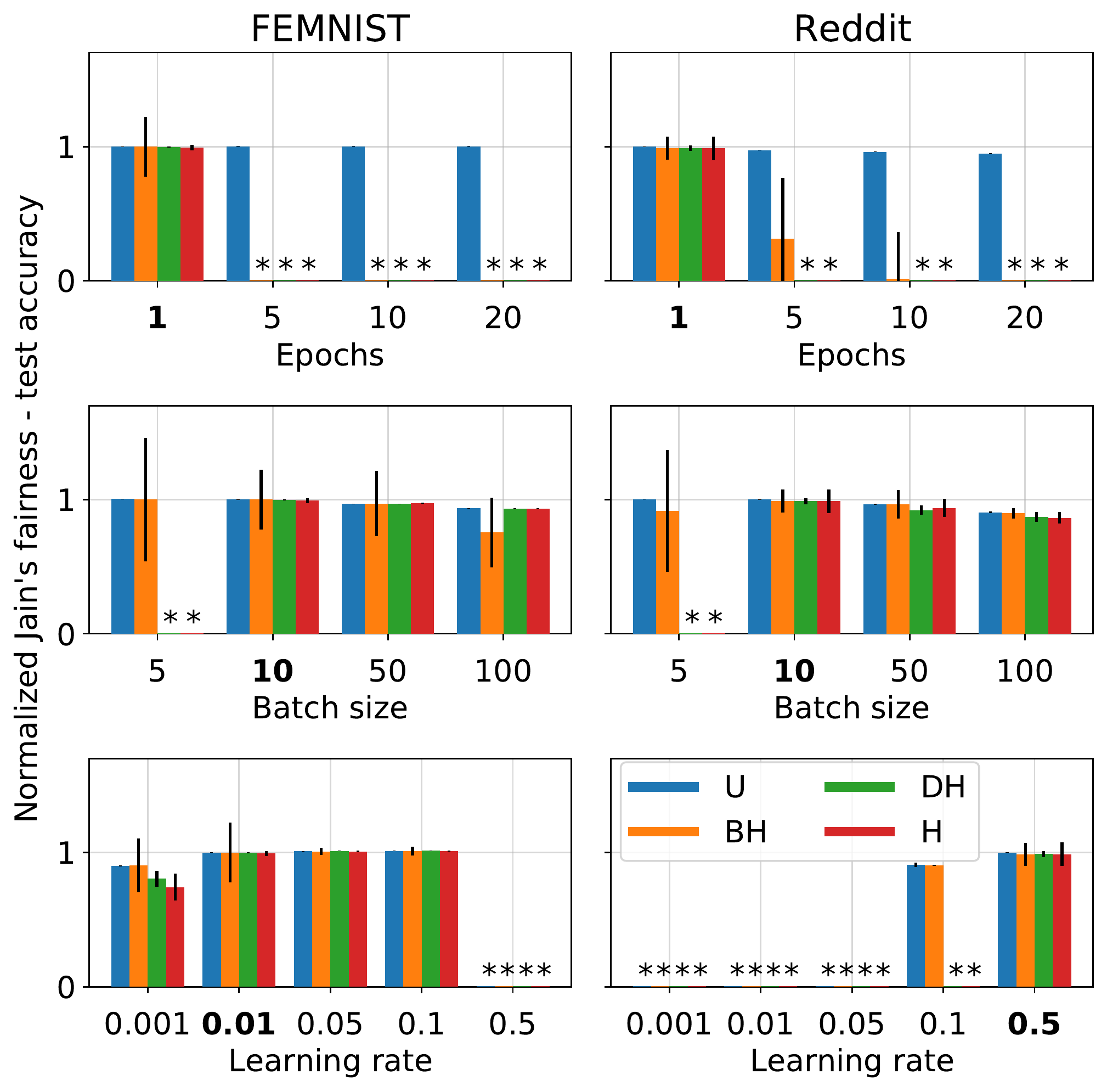}
   \caption{FEMNIST (Left), and Reddit (Right).}
  \end{subfigure}
  \\
    \begin{subfigure}{1\linewidth}
    \includegraphics[width=1\linewidth]{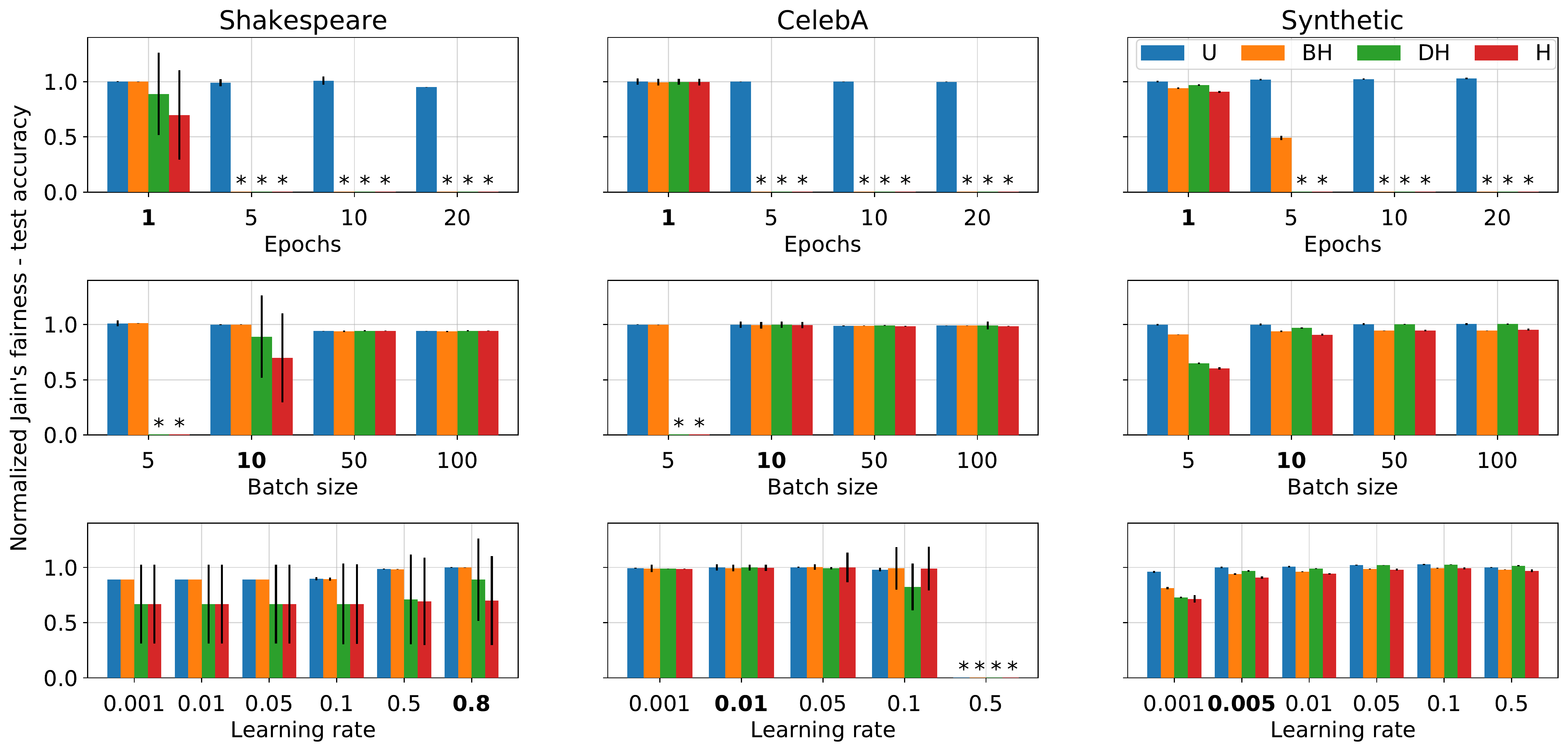}
    \caption{Shakespeare (Left), Celeb (Middle) and Synthetic (Right).}
    \end{subfigure}
    \caption{The sensitivity of normalized Jain's fairness of test accuracy to the choice of the application-specific hyper-parameter for various benchmarks. Epochs (Top), Batch size (Middle), and Learning rate (Bottom) rows.}
    \label{fig:hyper-fairness-test-acc-train-params}
\end{figure} 
\end{document}